\documentclass[10pt]{article}

\usepackage[utf8]{inputenc}
\usepackage[T1]{fontenc}
\usepackage{lmodern}
\usepackage[margin=1in]{geometry}
\usepackage{amsmath}
\usepackage{amssymb}
\usepackage{graphicx}
\usepackage{booktabs}
\usepackage[round]{natbib}
\usepackage{microtype}
\usepackage[colorlinks=true, linkcolor=blue!60!black, citecolor=blue!60!black, urlcolor=blue!60!black]{hyperref}
\usepackage{xcolor}

\graphicspath{{figures/}}

\newcommand{\known}{\textsc{known}}
\newcommand{\unknownreal}{\textsc{unknown-real}}
\newcommand{\fabricated}{\textsc{fabricated}}

\title{\textbf{Does Bielik Know What It Doesn't Know?\\
Activation Dispersion Separates Entity Familiarity\\
from Factual Reliability Across Model Scale}\thanks{Preprint. Code and data are available at the project repository: \url{https://github.com/agentGreg/bielik-hallucination-detection}.}}

\author{Grzegorz Brzezinka\\
Prosit AS\\
\texttt{greg@prosit.no}}

\date{July 2026}

\begin{document}

\maketitle

\begin{abstract}
\noindent
Large language models (LLMs) hallucinate most readily about entities they have never seen. We ask whether a model's activations betray entity familiarity before a single answer token is generated - and whether that signal says anything about the factual reliability of the answers. On four Polish Bielik models (1.5B--11B parameters), we probe four entity domains - athletes, cities, writers, and musicians - each with 42 well-known, 42 obscure-but-real, and 42 fabricated entities addressed by a one-sentence question (504 prompts per model). Two unsupervised, single-forward-pass dispersion measures over post-SwiGLU MLP activations - inverse participation ratio and spectral entropy - separate known from fabricated entities with AUROC 0.95--1.00 across all four domains and every scale; a supervised linear probe reaches 0.99--1.00. Both clear a selection-aware permutation floor of about 0.70--0.74 (95th percentile of a null that re-selects the best layer per permutation; empirical $p \le 10^{-3}$), survive held-out layer selection (0.93--0.99), and persist on real names: known versus real-but-obscure entities separate at 0.96--1.00, so fabricated-string artifacts are not the driver. The signal transfers across entity types: a probe trained on one domain and evaluated on another retains a mean off-diagonal AUROC of 0.92--0.99, with the only large drops occurring where the prompt template also changes (cities use a different question stem). A matched-template counterfactual - re-extracting cities and writers under one shared neutral stem - shows these drops are template-caused: transfer into cities recovers to 0.999--1.000 at every scale, leaving only a residual cities-as-source asymmetry. Per-head attention-entropy analysis shows the signal is diffuse rather than carried by a small fixed set of heads. This representational signal, taken as the better of the two metrics, is at ceiling on this contrast already at 1.5B - whether familiarity itself still improves with scale is not distinguishable at this ceiling. Behavioral factual reliability instead scales sharply: 0, 2, 10, and 19 of 42 known athletes are answered fully correctly by the 1.5B, 4.5B, 7B, and 11B models under a strict judge (6, 16, 24, 33 under a soft key-facts rubric). Within known entities, separating correct answers from hallucinations is much harder: the probe reaches 0.93, and dispersion metrics do no better than a first-token-entropy baseline. A five-sample semantic-entropy baseline attains 0.71--0.83 on the known-vs-fabricated contrast at five times the inference cost, though it wins the correctness contrast (up to 0.87) where dispersion fails. Despite this representational awareness, the models almost never hold back: an LLM audit of all 2{,}520 sampled athlete answers finds 2 refusals and 1 hedged answer (99.88\% direct assertions), both refusals from the largest model. Entity familiarity and factual reliability are distinct phenomena moving on different scaling curves. (Behavioral labels, semantic entropy, and the refusal audit are reported for athletes only; the other three domains carry condition-based detection labels.)
\end{abstract}

\section{Introduction}
\label{sec:intro}

When an LLM is asked about something it does not know, does its internal state look different from when it is on familiar ground? A physically motivated intuition says yes: knowledge retrieval should behave like a \emph{localized} excitation: a compact, selective activation pattern (memory retrieval), whereas confabulation should look \emph{delocalized}, with activation smeared broadly across the network. As a motivating analogy, no more than that, the dichotomy recalls localized versus extended eigenstates in disordered media \citep{anderson1958absence}, and it suggests cheap, unsupervised concentration statistics: the inverse participation ratio (IPR) and the entropy of the activation distribution, computed from a single forward pass with no labels, no sampling, and no auxiliary model.

The literature warns that the naive version of this intuition is almost certainly false. First, large language models are always strongly \emph{contextually sparse}: for any given token, most attention heads and most Multi-Layer Perceptron (MLP) parameters can be silenced with little effect on the output, regardless of whether the model knows the answer \citep{liu2023dejavu,li2023lazy}. ``The whole network lights up uniformly'' is not an operating regime that actually occurs; if there is a signal, it must be a quantitative and geometric difference in \emph{how} concentrated the activity is, not a qualitative one. Second, mechanistic work locates knowledge awareness in \emph{low-dimensional directions} rather than in global activation volume: sparse-autoencoder features that fire on known versus unknown entities causally steer refusal and hallucination behavior \citep{ferrando2025entity}, and circuit-tracing analyses find ``known answer'' features whose misfiring produces hallucinations \citep{lindsey2025biology}. A global dispersion measure could easily average such a directional signal into noise.

This leaves a concrete, if narrower than it once was, empirical gap. Unsupervised single-pass dispersion statistics for hallucination detection have recently appeared: D\textsuperscript{2}HScore computes intra-layer dispersion of token representations plus inter-layer drift \citep{ding2025d2hscore}, EigenTrack tracks covariance-spectrum statistics of hidden states, including spectral entropy, during generation \citep{ettori2025eigentrack}, and MIND builds an unsupervised internal-state detector from pseudo-labeled generations \citep{su2024mind}. All of these, however, score \emph{generated answers} using token-span or covariance-level quantities. What remains untested, to our knowledge, is \emph{closed-form, per-neuron dispersion of a single activation vector} (IPR and per-neuron spectral entropy - the physics-flavored localization statistics) read \emph{at the prompt point, before any answer token}, as a detector of \emph{knowledge presence} rather than answer-level hallucination - benchmarked simultaneously against the supervised ceiling (a linear probe), a selection-aware permutation floor, a token-count lexical baseline, and the standard multi-sample baseline (semantic entropy), across a scale sweep. Nor has any internal-state hallucination study been carried out on a Polish model family: the multilingual shared task Mu-SHROOM spans 14 languages without Polish \citep{vazquez2025mushroom}, and existing multilingual internal-state probing covers French, Bangla, and Amharic \citep{alvi2026multihaludet} - while tokenization of Polish named entities and knowledge of Polish domains are exactly the conditions under which a Polish-first model like Bielik \citep{ociepa2025bielikv3small,ociepa2025bielik11bv2} is deployed.

We run this test across four Bielik v3.0 instruction-tuned models (1.5B, 4.5B, an 11B, and a 7B Minitron variant obtained by pruning and distillation), on a three-condition dataset spanning four entity domains: Polish athletes, cities, writers, and musicians. Each condition is holding famous (\known), real but obscure (\unknownreal), and fabricated but morphologically plausible (\fabricated) entities, with fabricated names approximately length-matched to famous ones (the residual mismatch is measured and reported as an explicit per-domain lexical baseline). The four domains let us test whether the familiarity signal is a property of a single entity type or generalizes across them \citep[cf.][]{ferrando2025entity}. Beyond detection, we measure (on the athletes domain) the model's \emph{behavioral} reliability on the same entities: how often it actually answers correctly about entities it plausibly knows, which lets us put the representational signal and the behavioral phenotype on the same scale axis.

\paragraph{Contributions.}
\begin{enumerate}
  \item \textbf{A systematic test of closed-form, per-neuron activation dispersion at the prompt point as an unsupervised knowledge-presence detector, with full uncertainty quantification.} One-pass IPR and spectral entropy over post-SwiGLU MLP activations separate \known{} from \fabricated{} athletes with Area Under the Receiver Operating Characteristic (AUROC) 0.94--0.98 on every model in the sweep; a supervised linear probe reaches 0.99--1.00. All values clear selection-aware permutation floors of $\approx$0.70--0.74 (empirical $p \le 10^{-3}$) and survive held-out layer selection at 0.93--0.99 (Section~\ref{sec:dispersion}). On the fabricated-string-free control contrast (\known{} vs.\ \unknownreal), dispersion reaches 0.97--1.00 (Section~\ref{sec:controls}).
  \item \textbf{Generalization across four entity types.} The same one-pass detector separates \known{} from \fabricated{} across athletes, cities, writers, and musicians (best dispersion 0.95--1.00, probe 0.99--1.00), and probes transfer across domains at mean off-diagonal AUROC 0.987/0.959/0.919/0.931 (1.5B/4.5B/7B/11B). The only large transfer drops involve cities, the one domain whose prompt template differs; a matched-template counterfactual shows the target-side drop is template-caused. Under a shared neutral stem, transfer into cities recovers to 0.999--1.000 at every scale, with a residual cities-as-source asymmetry (0.82--0.98) consistent with a tighter places manifold. A per-head analysis shows the attention-entropy signal is diffuse, with only 1--2 heads shared across all four domains' top-20 (Section~\ref{sec:generalization}).
  \item \textbf{Two dissociated scaling curves, measured on one dataset.} The representational familiarity signal (best of the metric pair) is at ceiling on this contrast already at 1.5B, while behavioral factual reliability rises sharply with scale: 0$\rightarrow$2$\rightarrow$10$\rightarrow$19 of 42 \known{} athletes answered fully correctly under a strict judge, 6$\rightarrow$16$\rightarrow$24$\rightarrow$33 under a soft key-facts rubric (Section~\ref{sec:scaling}). This ordering is what frequency-based accounts of factual recall predict \citep{mallen2023trust,kandpal2023longtail}; the contribution is the clean same-dataset, same-entities measurement of both axes across scale.
  \item \textbf{An honest mostly-negative result on the behavioral axis.} Within \known{} athletes, correct-vs-hallucination is far harder than known-vs-fabricated, and the full model $\times$ threshold grid is inconsistent: dispersion sometimes beats the probe, the probe once falls below its own permutation floor, and the strongest cell (11B probe 0.929) is suggestive rather than established. No metric we tested reads factual reliability the way dispersion reads familiarity (Section~\ref{sec:behavioral}).
  \item \textbf{A complementarity analysis against semantic entropy.} Discrete semantic entropy over five samples reaches AUROC 0.71--0.83 on known-vs-fabricated - below the one-pass dispersion signal with non-overlapping CIs - but wins the correctness contrast (up to 0.87) where dispersion fails; the two consume different information (prompt-side familiarity vs.\ generation-side consistency) (Section~\ref{sec:se}).
  \item \textbf{A near-zero-refusal regime, quantified by an answer-level LLM audit.} Across 2{,}520 sampled athlete answers (42 entities $\times$ 3 conditions $\times$ 5 samples $\times$ 4 models), the models decline twice and hedge once (99.88\% direct assertions). Only the largest model ever abstains - despite carrying a near-perfectly decodable internal familiarity signal at every scale. This is a concrete, measurable alignment gap for Polish LLM deployments (Section~\ref{sec:discussion}).
\end{enumerate}

\section{Related work}
\label{sec:related}

\paragraph{White-box detectors on internal states.}
\citet{azaria2023internal} first showed that a classifier on hidden activations predicts statement truthfulness, establishing the supervised-probe paradigm; \citet{orgad2025llmsknow} sharpened it, finding truthfulness information concentrated in exact-answer tokens, and notably probes that fail to generalize across datasets. Unsupervised and dispersion-flavored detectors are an active line. INSIDE \citep{chen2024inside} performs eigen-analysis of the covariance of hidden states across multiple sampled responses (EigenScore); LLM-Check \citep{sriramanan2024llmcheck} makes the covariance line single-pass - its Hidden Score is the log-determinant of the Gram matrix of hidden states over response tokens, alongside logit-entropy and attention-kernel scores; D\textsuperscript{2}HScore \citep{ding2025d2hscore} scores hallucination from training-free intra-layer dispersion of token representations plus inter-layer drift; EigenTrack \citep{ettori2025eigentrack} tracks covariance-spectrum statistics, including spectral entropy, during generation; MIND \citep{su2024mind} trains an internal-state detector without human labels from pseudo-labeled generations; HaloScope \citep{du2024haloscope} finds a truth subspace in activation space from unlabeled generations. Semantic Entropy Probes \citep{kossen2024semantic} distill the multi-sample semantic-entropy signal into a cheap linear probe on hidden states; Lookback Lens \citep{chuang2024lookback} detects contextual hallucination from attention-allocation ratios alone; \citet{binkowski2025lapeigvals} use spectral features (Laplacian eigenvalues) of attention maps; TSV \citep{park2025tsv} steers latents with learned truthful directions. All of the unsupervised detectors above score \emph{generated answers} from token-span or covariance-level statistics. Our contribution is deliberately minimalist and differently aimed: two closed-form \emph{per-neuron} dispersion statistics of a \emph{single} activation vector, read at the \emph{prompt point} before generation, as a detector of \emph{knowledge presence} - no training, one forward pass - evaluated against the probe that upper-bounds what the activations contain.

\paragraph{Black-box baselines.}
Semantic entropy \citep{kuhn2023semantic,farquhar2024detecting} clusters multiple sampled answers by bidirectional entailment and measures entropy over meaning classes; SelfCheckGPT \citep{manakul2023selfcheckgpt} checks sampled answers for mutual consistency. These methods are model-agnostic but require several forward passes plus auxiliary inference; they are our cost reference point (Section~\ref{sec:se}).

\paragraph{Self-knowledge, long-tail knowledge, and mechanistic accounts.}
\citet{kadavath2022languagemodels} showed that models can largely predict whether they know the answer to a question ($P(\mathrm{IK})$), and \citet{yin2023selfaware} benchmarked whether models know what they do not know at the behavioral level - both foreshadowing the familiarity/reliability split we measure internally and externally on one dataset. Popularity and pretraining-frequency effects bear directly on that split: factual recall tracks entity popularity \citep{mallen2023trust} and models struggle to learn long-tail knowledge without many supporting documents \citep{kandpal2023longtail}, which predicts exactly the pattern we observe: entity recognition is arriving long before reliable facts about the entity. Mechanistically, \citet{ferrando2025entity} identified causal known-entity/unknown-entity SAE directions on the entity token, which modulate attribute-extraction attention heads; \citet{obeso2025realtime} carried that entity-familiarity signal into real-time detection of hallucinated entities in long-form generation; \citet{lindsey2025biology} traced known-answer circuits whose failure modes produce hallucination; and \citet{wu2025retrievalheads} showed that long-context factuality is carried by a sparse ($<$5\%), causally necessary set of retrieval heads. Closest to our probe, \citet{cencerrado2025noanswer} predict answer accuracy from \emph{question-only} linear probes - supervised pre-generation prediction, evaluated against answer-level correctness across 7B--70B models; our supervised probe replicates that pre-generation readability on a new family and language, while our headline detector needs no labels at all. Our dispersion results are a coarse-grained, label-free shadow of this machinery.

\paragraph{Non-English and multilingual hallucination detection.}
The multilingual state of the art bounds our language claim. Mu-SHROOM, the SemEval-2025 shared task on multilingual hallucination detection, spans 14 languages and 43 teams - without Polish - and operates at the output-span level \citep{vazquez2025mushroom}. MultiHaluDet probes hidden states multilingually, covering French, Bangla, and Amharic \citep{alvi2026multihaludet}; HalluVerse25 offers a fine-grained multilingual benchmark (English, Arabic, Turkish) with no internal-state analysis \citep{abdaljalil2025halluverse}. We find no internal-state hallucination study on a Polish-native model family; to our knowledge this work is the first.

\paragraph{Pitfalls for global measures.}
Massive activations - a handful of neurons with values orders of magnitude larger than the rest, largely input-independent \citep{sun2024massive} - can dominate any participation-ratio statistic; we winsorize activations before computing dispersion (Section~\ref{sec:metrics}). The effective rank of a representation matrix \citep{roy2007effective} extends dispersion from single vectors to token spans and is a one-pass relative of EigenScore; we include it in the extended metric set. Contextual-sparsity results \citep{liu2023dejavu,li2023lazy} calibrate expectations: sparsity is ubiquitous and input-dependent, so only \emph{relative} dispersion differences between conditions are meaningful.

\section{Method}
\label{sec:method}

\subsection{Dataset: three conditions across four entity domains}
\label{sec:dataset}

We construct the dataset in four entity domains: athletes, cities, writers, and musicians, each holding three conditions of 42 entities,\footnote{The count of 42 per condition is a deliberate nod to \citet{adams1979hitchhiker}.} for 126 entities per domain and 504 per model:
\begin{itemize}
  \item \known: famous entities the models can be presumed to have seen abundantly in pretraining (e.g., Robert Lewandowski and Adam Małysz for athletes, Warszawa for cities, Adam Mickiewicz for writers, Fryderyk Chopin for musicians);
  \item \unknownreal: real but obscure entities (lower-league athletes, small villages, minor authors and performers) that the models almost certainly cannot describe;
  \item \fabricated: invented names, morphologically well-formed in Polish (in the spirit of ``Roman Lewandowicz''), constructed to be approximately length-matched to the \known{} set of the same domain.
\end{itemize}
The athletes domain is our primary domain: it alone carries behavioral labels, semantic-entropy scores, and the refusal audit (Sections~\ref{sec:labeling}--\ref{sec:se-method}); the other three domains carry condition-based detection labels only. The athletes set extends the original 41-per-condition list of the round-1 study by one entity per condition.

Two Polish prompt templates are used, differing only in the question stem so that a person is asked \emph{who} and a place \emph{what}:
\begin{quote}
  \emph{Kim jest \{entity\}? Odpowiedz jednym zdaniem.} (``Who is \{entity\}? Answer in one sentence.'')  -  athletes, writers, musicians;\\
  \emph{Czym jest \{entity\}? Odpowiedz jednym zdaniem.} (``What is \{entity\}? Answer in one sentence.'')  -  cities.
\end{quote}
Within a domain the template is uniform, so all variation between prompts is confined to the entity span; the stem difference between the cities template and the three people templates is a controlled confound we return to in Section~\ref{sec:generalization}. For the template-causality control reported there, the cities and writers domains were additionally re-extracted under a third, neutral stem grammatical for both people and places - \emph{Co wiesz o \{entity\}? Odpowiedz jednym zdaniem.} (``What do you know about \{entity\}? Answer in one sentence.''), on all four models.

The length matching is only approximate, and the model family uses \emph{two} tokenizers, not one: 1.5B and 4.5B share a tokenizer under which athlete names span 2--8 tokens (\known{} mean 3.19, \unknownreal{} 3.98, \fabricated{} 3.90), while Minitron-7B and 11B share a different one under which the same names span 4--11 tokens (means 6.17, 6.02, 6.76). Entity token count alone therefore separates \known{} from \fabricated{} athletes at AUROC 0.681 (small-model tokenizer) and 0.612 (large-model tokenizer), a weak but nonzero lexical baseline that we report explicitly and control for in Section~\ref{sec:controls}. It is however far below the dispersion signal. The three new domains were matched at least as tightly as the athletes baseline: token-count-only K-vs-F AUROC is 0.637/0.494 (cities), 0.500/0.500 (writers), and 0.509/0.505 (musicians) under the small/large tokenizers - writers and musicians at chance, cities saturating at 0.637 under the small tokenizer because a handful of single-token major cities cannot be matched by any invented multi-token toponym. Full per-domain screening, replacement logs, and web-verification coverage are documented with the release. The experiments were orchestrated with AI coding assistance. All results were produced and verified by the author.

\subsection{Behavioral labeling (athletes)}
\label{sec:labeling}

Behavioral labeling is carried out on the athletes domain only. For each entity and model we sample five answers (chat template, at most 64 new tokens each). Answers to \known{} entities are graded by an LLM judge (Claude Opus 4.8) in two regimes:
\begin{itemize}
  \item \textbf{Strict}: a deliberately minimal judge - a bare one-line prompt asking whether the answer is factually correct about the real world, with no rubric, no tolerance for answers truncated by the 64-token generation cap, and unparseable judge outputs defaulting to \emph{incorrect}. An entity is \emph{fully correct} if all 5 of 5 samples pass. Small models fail this bar mostly by decorating true identities with invented achievements (e.g., a world-championship title Lewandowski never won), but a part of the strict failure count reflects truncation and parse defaults rather than confabulation - one motivation for the soft rubric below.
  \item \textbf{Soft}: a key-facts rubric - identity and discipline must be correct; fabricated \emph{significant} achievements still count as errors; harmless omissions and truncations are tolerated. We report thresholds of 5/5, $\geq$4/5, and $\geq$3/5 correct samples.
\end{itemize}
The main detection contrast (\known{} vs.\ \fabricated) is \emph{condition-based} and therefore independent of the judge; judges enter only through the behavioral axis (Section~\ref{sec:behavioral}). In manual spot checks of 615 soft-judge verdicts (5 samples $\times$ 41 \known{} entities $\times$ three dense models, on the round-1 athletes list) we found 2 anomalous verdicts ($\approx$0.3\%); see Section~\ref{sec:limitations}.

Refusal and hedging are audited at the \emph{answer} level over all three athlete conditions: an LLM classifier (Claude Haiku 4.5) with randomized human control labels each of the 2{,}520 stored answers (42 entities $\times$ 3 conditions $\times$ 5 samples $\times$ 4 models) as a direct assertion, a refusal, or a hedged non-answer (Section~\ref{sec:discussion}). An earlier six-substring marker heuristic proved brittle in both directions: its single hit was a false positive on a song title, and it missed both refusals the audit found, so we report only the LLM-audited counts.

\subsection{Models}
\label{sec:models}

We sweep four instruction-tuned Bielik v3.0 checkpoints \citep{ociepa2025bielikv3small,ociepa2025bielik11bv2}: \textbf{1.5B} (32 transformer layers, $d=1536$), \textbf{4.5B} (60 layers, $d=2048$), \textbf{11B} (50 layers, $d=4096$), and \textbf{Minitron-7B} (40 layers, $d=4096$) \citep{kinas2026minitron}, the last obtained by Minitron-style pruning and knowledge distillation \citep{muralidharan2024minitron} from a larger Bielik model rather than trained densely from scratch. The Minitron point therefore extends the scale curve between 4.5B and 11B but is architecturally not a pure member of the dense family; we flag it explicitly wherever it behaves differently. All models share a Mistral-like decoder architecture with SwiGLU MLPs.

\subsection{Signal extraction and measurement points}
\label{sec:extraction}

For every prompt we run a single forward pass and record, at each layer: (i) the post-SwiGLU gate activations of the MLP (via forward hooks on \texttt{act\_fn}); (ii) residual-stream hidden states; (iii) attention maps; and (iv) logit-lens distributions (hidden state decoded through the output head). Signals are read at two token positions:
\begin{itemize}
  \item the \textbf{entity point}: the last token of the entity name, where mechanistic work locates entity-recognition features \citep{ferrando2025entity};
  \item the \textbf{prompt point}: the last token of the prompt, several tokens downstream of the entity.
\end{itemize}
The entity point yields nominally higher separability, but it is also maximally exposed to a lexical confound: a rare fabricated string can produce atypical activation statistics for reasons of token frequency alone, independent of knowledge. The prompt point, where the model has integrated the question and lexical identity is no longer the input, is the more conservative measurement, and \emph{all headline numbers in this paper are prompt-point numbers}. Entity-token indices are computed from the very encoding fed to the model, avoiding Beginning of Sequence (BOS) off-by-one misalignment (verified empirically).

\subsection{Dispersion metrics}
\label{sec:metrics}

Let $a \in \mathbb{R}^{d_\mathrm{mlp}}$ be the vector of post-SwiGLU MLP activations of layer $\ell$ at the measurement token. Define the normalized weights
\begin{equation}
  p_i = \frac{a_i^2}{\sum_j a_j^2},
\end{equation}
and from them
\begin{equation}
  \mathrm{IPR}(\ell) = \sum_i p_i^2,
  \qquad
  S(\ell) = -\sum_i p_i \log p_i .
\end{equation}
$1/\mathrm{IPR}$ is the effective number of active neurons; $S$ is the Shannon entropy of the activation distribution (labelled \emph{spectral entropy} in the figures). Localized (concentrated) activity means high Inverse Participation Ration (IPR) and low $S$; delocalized activity the reverse. Because a few massive-activation outliers can dominate these statistics irrespective of content \citep{sun2024massive}, activations are winsorized (nearest-value clipping above the $q=0.99$ quantile) before computing $p$. Since SwiGLU activations have no exact zeros \citep{li2023lazy}, both measures are continuous rather than combinatorial, which is irrelevant for IPR and entropy.

The winsorization quantile is a researcher degree of freedom sitting inside the headline metric, so we probed it directly on the 1.5B model by re-extracting raw activations and sweeping $q \in \{0.95, 0.99, 0.999, 1.0\}$ (the last effectively skips winsorization). A strong signal exists at every setting - best-layer AUROC 0.899/0.933/0.836/0.889 for IPR and 0.910/0.944/0.890/0.905 for entropy - but the paper's setting $q=0.99$ is the most favorable of the four, the best layer moves with $q$, and the per-layer value at the layer reported in Table~\ref{tab:headline} degrades substantially at other settings (entropy at layer 22: 0.509 at $q=0.95$, 0.782 at $q=0.999$, 0.731 without winsorization). Dispersion-as-familiarity is robust to the choice, while specific layer-level values are not. Only the 1.5B model was swept.

The extended metric set comprises of: the \textbf{effective rank} \citep{roy2007effective} of the matrix of hidden states over the prompt tokens (per layer) - the exponential of the entropy of the normalized singular-value spectrum, a one-pass relative of EigenScore \citep{chen2024inside}; \textbf{attention entropy} per head, averaged per layer, in raw and length-normalized variants; and \textbf{logit-lens entropy} \citep{nostalgebraist2020logitlens} per layer, measuring how early in depth the model commits to an output distribution. As a strictly cheaper output-side baseline we use the \textbf{entropy of the first answer-token logprobs}.

\subsection{Probe, controls, and evaluation}
\label{sec:probe}

For each residual layer we train an $\ell_2$-regularized logistic-regression probe on the raw hidden state (5-fold cross-validated), giving a per-layer supervised \emph{upper bound} on the linearly decodable signal.

Because every headline number is a maximum over layers (and, for dispersion, over the \{IPR, entropy\} pair), we calibrate selection optimism explicitly with three complementary tools. \textbf{Permutation null with selection:} we permute condition labels and rerun the \emph{entire} pipeline, including the layer (and metric) re-selection, per permutation - 10{,}000 permutations for dispersion, 1{,}000 for the probe - and take the 95th percentile of the null maxima as the floor. These selection-aware floors land at 0.701--0.715 (dispersion) and 0.721--0.742 (probe) across the four models, with null means around 0.65; a single label permutation, by contrast, yields 0.62--0.64 and understates the null's upper range. Any metric must clear these floors, not 0.5. \textbf{Bootstrap confidence intervals (CI):} 95\% CIs from 10{,}000 stratified resamples with the layer fixed at the observed best layer; these quantify sampling noise, not selection - and because they condition on the observed best layer, they are mildly optimistic post-selection (the held-out analysis below is the out-of-sample complement). \textbf{Held-out layer selection:} 200 stratified half-splits in which the layer (and metric) is selected on one half of the entities and the AUROC is evaluated on the other half.

All discrimination results are AUROC, reported as separability $\max(\mathrm{AUC},\, 1-\mathrm{AUC})$; the primary contrast is \known{} (0) vs.\ \fabricated{} (1) with $n = 84$ per contrast (42 per class; likewise 84 per contrast in each of the three new domains).

\subsection{Semantic-entropy baseline}
\label{sec:se-method}

From the five stored samples per athlete entity we compute discrete semantic entropy \citep{kuhn2023semantic,farquhar2024detecting}. The answers are clustered into meaning-equivalence classes by an LLM clustering judge (Claude Sonnet 5, bidirectional-entailment criterion), and the entropy of the resulting cluster distribution is the score. Clustering succeeded for every row (zero fallback rows in all $4\times126$ athlete entities). This baseline costs five forward passes of the subject model plus clustering calls, versus one forward pass and no auxiliary model for dispersion.

\section{Results}
\label{sec:results}

\subsection{Dispersion separates known from fabricated at every scale}
\label{sec:dispersion}

\begin{figure}[!tbp]
\centering
\includegraphics[width=0.95\textwidth]{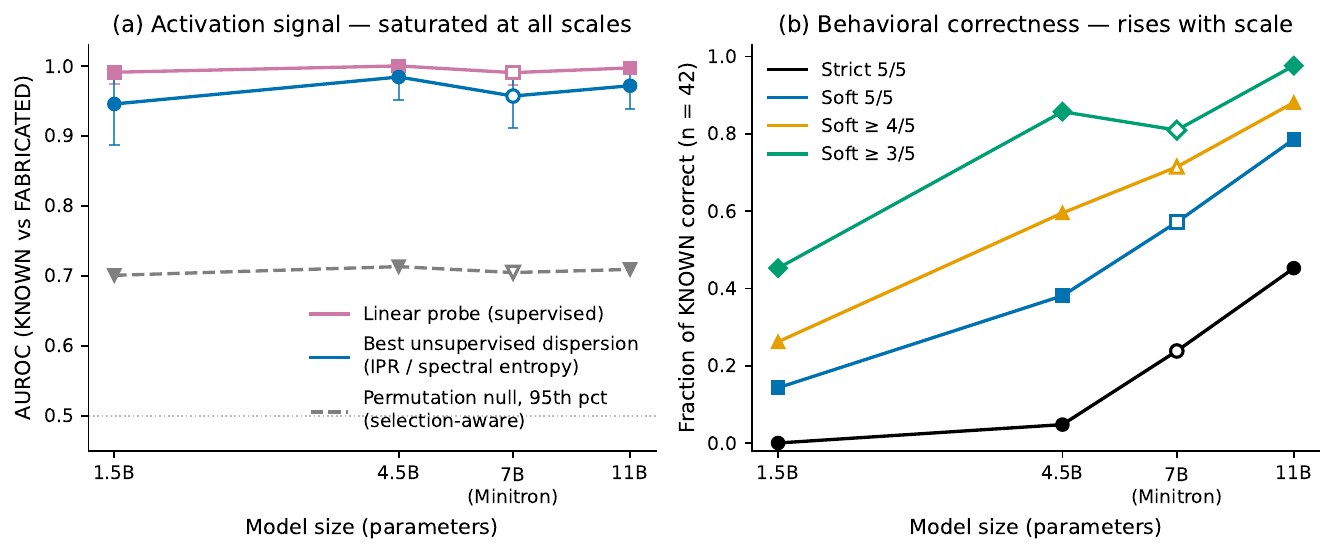}
\caption{Two scaling curves that do not move together (athletes domain). \textbf{(a)} The activation-level familiarity signal (\known{} vs.\ \fabricated, prompt point) is at or near ceiling on this contrast at every scale: unsupervised dispersion 0.94--0.98, supervised probe 0.99--1.00. Error bars: 95\% bootstrap CIs at the fixed best layer. Dashed line: selection-aware floor - the 95th percentile of a permutation null with layer selection ($\approx$0.70--0.71). \textbf{(b)} Behavioral correctness on the same \known{} entities rises steeply with scale under every judging threshold; the strict criterion goes from 0/42 at 1.5B to 19/42 at 11B. Open markers denote the Minitron-7B point, a pruned/distilled variant rather than a dense family member. Even the smallest model's representations encode which entities it knows; what scales is whether it can answer about them correctly.}
\label{fig:scale}
\end{figure}

Table~\ref{tab:headline} and Figure~\ref{fig:scale} (left) give the headline result for athletes. On every model in the sweep, at least one unsupervised dispersion metric separates \known{} from \fabricated{} with AUROC between 0.946 [0.887, 0.988] (1.5B) and 0.984 [0.952, 1.000] (4.5B), read from a single forward pass at the prompt point with no labels. The supervised probe confirms that the information content of the activations is near-ceiling everywhere (0.990--1.000).

Neither number is an artifact of layer-maximum selection (Table~\ref{tab:robust}). In 10{,}000 label permutations that re-select the best layer and metric, the null maximum never reached the observed dispersion values (95th percentiles 0.701--0.715; empirical $p \le 10^{-4}$); the probe behaves identically against its own null (95th percentiles 0.721--0.742; $p \le 10^{-3}$ at 1{,}000 permutations). The signal also survives held-out selection: choosing the layer (and metric) on a random half of the entities and evaluating on the other half yields out-of-sample dispersion AUROCs of 0.928--0.970 and probe AUROCs of 0.977--0.993. The unsupervised metrics recover most of the gap between the selection floor and the probe ceiling, so the hypothesis in its \emph{volumetric} form holds: the global dispersion measures work, not only low-dimensional directions. Section~\ref{sec:generalization} shows the same detector holds across three further entity domains.

\begin{table}[!tbp]
\centering
\footnotesize
\caption{Headline results (athletes domain): \known{} vs.\ \fabricated{} separability (AUROC, prompt point; $n=84$), with best layer in parentheses and 95\% bootstrap CIs (10{,}000 stratified resamples, layer fixed at the observed best layer). Dispersion metrics are unsupervised, single forward pass; the probe is the supervised upper bound (5-fold CV). \textbf{Bold} marks the better of the two unsupervised metrics per model; the same convention is used in all tables. Behavioral correctness counts for the same models appear in Figure~\ref{fig:scale}(b) and Section~\ref{sec:scaling}; the cross-domain version of this contrast is in Section~\ref{sec:generalization}.}
\label{tab:headline}
\begin{tabular}{lccc}
\toprule
Model & IPR (layer) & Spectral entropy (layer) & Probe (layer) \\
\midrule
1.5B & 0.936 (22) [0.877, 0.981] & \textbf{0.946} (22) [0.887, 0.988] & 0.991 (20) [0.974, 1.000] \\
4.5B & 0.984 (50) [0.946, 1.000] & \textbf{0.984} (36) [0.952, 1.000] & 1.000 (36) [1.000, 1.000] \\
7B (Minitron) & 0.914 (17) [0.844, 0.968] & \textbf{0.957} (26) [0.912, 0.990] & 0.990 (27) [0.973, 1.000] \\
11B & \textbf{0.972} (35) [0.939, 0.994] & 0.896 (35) [0.827, 0.954] & 0.997 (31) [0.989, 1.000] \\
\bottomrule
\end{tabular}
\end{table}

\begin{table}[!tbp]
\centering
\footnotesize
\caption{Selection-robustness of the headline (\known{} vs.\ \fabricated, prompt point). Held-out: mean $\pm$ SD over 200 stratified half-splits with the layer (and, for dispersion, the metric) selected on one half and AUROC evaluated on the other. Floors: 95th percentile of the permutation null that re-selects the best layer (and metric) per permutation - 10{,}000 permutations for dispersion, 1{,}000 for the probe; no permutation reached the observed values (empirical $p \le 10^{-4}$ and $\le 10^{-3}$ respectively).}
\label{tab:robust}
\begin{tabular}{lcccc}
\toprule
& \multicolumn{2}{c}{Held-out layer selection} & \multicolumn{2}{c}{Permutation floor (null 95th pct)} \\
\cmidrule(lr){2-3}\cmidrule(lr){4-5}
Model & Dispersion & Probe & Dispersion & Probe \\
\midrule
1.5B & $0.928 \pm 0.039$ & $0.977 \pm 0.018$ & 0.701 & 0.726 \\
4.5B & $0.968 \pm 0.030$ & $0.993 \pm 0.011$ & 0.713 & 0.742 \\
7B (Minitron) & $0.940 \pm 0.041$ & $0.977 \pm 0.020$ & 0.705 & 0.721 \\
11B & $0.970 \pm 0.025$ & $0.991 \pm 0.014$ & 0.709 & 0.730 \\
\bottomrule
\end{tabular}
\end{table}

\subsection{Control contrasts: the signal is familiarity, with a measurable lexical residue}
\label{sec:controls}

\begin{table}[!tbp]
\centering
\footnotesize
\caption{Control contrasts (prompt point; best unsupervised dispersion metric with layer and 95\% bootstrap CI). \known{} vs.\ \unknownreal{} involves only real names with natural subword statistics and is therefore immune to fabricated-string artifacts; \unknownreal{} vs.\ \fabricated{} contrasts two conditions the model is (approximately) equally unfamiliar with, upper-bounding the lexical/plausibility component. The last column is the token-count-only lexical baseline for the headline contrast.}
\label{tab:controls}
\begin{tabular}{lccc}
\toprule
Model & K vs.\ UR (dispersion) & UR vs.\ F (dispersion) & Token count, K vs.\ F \\
\midrule
1.5B & 0.989 (23) [0.970, 1.000] & 0.699 (23) [0.585, 0.808] & 0.681 [0.572, 0.783] \\
4.5B & 0.994 (40) [0.982, 1.000] & 0.741 (39) [0.630, 0.843] & 0.681 [0.571, 0.781] \\
7B (Minitron) & 1.000 (17) [1.000, 1.000] & 0.824 (15) [0.728, 0.906] & 0.612 [0.491, 0.728] \\
11B & 0.971 (24) [0.936, 0.993] & 0.806 (47) [0.701, 0.897] & 0.612 [0.488, 0.727] \\
\bottomrule
\end{tabular}
\end{table}

The three-condition design supports two control contrasts that bound how much of the headline could be lexical rather than epistemic (Table~\ref{tab:controls}).

\textbf{Known vs.\ unknown-real (real names only).} Dispersion separates famous from obscure \emph{real} athletes at 0.971--1.000 - as strong as or stronger than the headline contrast (probe: 0.998--1.000) - on names with natural Polish subword statistics, where the token-count baseline is 0.694 under the small-model tokenizer and near chance (0.518) under the large-model one. This is the strongest single piece of evidence that the signal tracks familiarity rather than fabricated-string weirdness.

\textbf{Unknown-real vs.\ fabricated (both unfamiliar).} If dispersion were a pure familiarity readout, this contrast should sit near 0.5. But it does not, the dispersion reaches 0.699--0.824 (probe 0.819--0.913), while token count alone is at chance for the small-model tokenizer (0.515) and 0.638 for the large-model one. Some residual property of the fabricated strings (lexical statistics beyond raw length, or morphological plausibility) is detectable in the activations. We treat 0.70--0.82 as an \emph{upper bound} on the lexical/plausibility component of the headline contrast (it may also reflect genuine partial familiarity with real name-forms): large enough that the headline should not be read as purely epistemic, far too small to account for 0.94--0.98, and cleanly excluded from the \known{}-vs-\unknownreal{} control above.

\subsection{Two scaling curves: signal at ceiling, rising reliability}
\label{sec:scaling}

Figure~\ref{fig:scale} juxtaposes the two axes for athletes. The activation signal (panel a) is flat and at or near ceiling \emph{for the best-of-pair statistic} $\max(\mathrm{IPR}, S)$: it peaks at 4.5B (0.984 [0.952, 1.000]) and dips at 11B (0.972 [0.939, 0.994]); with bootstrap CIs spanning roughly $\pm$0.03--0.06, the dip is uninterpretable as a trend, so we read the statistic as at ceiling on this contrast at every scale - noting that famous-vs-fabricated is an easy contrast, and a ceiling here does not rule out further growth of familiarity itself on a graded-popularity axis. The individual metrics are less stable than the pair: the better metric flips from entropy (1.5B, 4.5B, 7B) to IPR (11B), and 11B entropy falls to 0.896 [0.827, 0.954], whose CI barely touches the 4.5B entropy CI [0.952, 1.000]; held-out selection shows the same picture (11B entropy $0.836 \pm 0.056$ out-of-sample vs.\ IPR $0.972 \pm 0.016$). The claim that survives scrutiny is that \emph{some} member of the dispersion pair is near ceiling on this contrast at every scale - not that either metric alone is scale-stable. Behavioral reliability (panel b) tells the opposite story: fully correct \known{} answers go 0$\rightarrow$2$\rightarrow$10$\rightarrow$19 of 42 (strict) and 6$\rightarrow$16$\rightarrow$24$\rightarrow$33 (soft 5/5) from 1.5B to 11B. Small models know \emph{which} entities are familiar - their activations say so almost perfectly - yet routinely decorate answers about those entities with fabricated details.

The Minitron-7B point fits the behavioral curve cleanly between 4.5B and 11B on the strict and soft-5/5 criteria (10 and 24 of 42), which is itself noteworthy: pruning and distillation preserved both the familiarity geometry (dispersion 0.957) and a scale-appropriate level of factual reliability. The one exception is the most lenient threshold, where Minitron is non-monotonic (soft $\geq$3/5: 34 vs.\ 36 for the 4.5B model) - a two-entity difference we do not over-interpret here.

\subsection{Depth structure of the signal}
\label{sec:depth}

\begin{figure}[!tbp]
\centering
\includegraphics[width=0.92\textwidth]{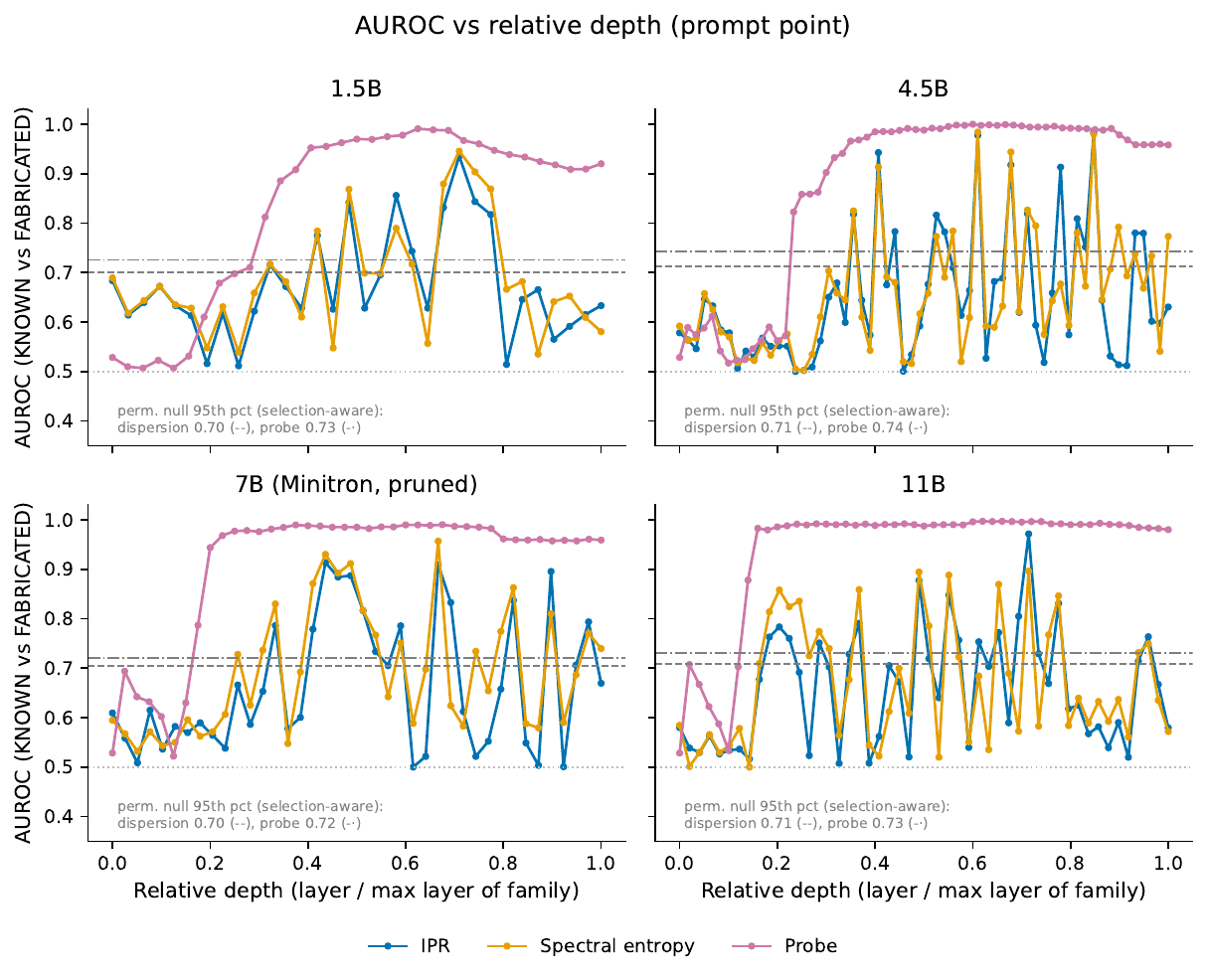}
\caption{AUROC (\known{} vs.\ \fabricated, prompt point) versus \emph{relative} depth (layer index normalized by the deepest layer of its family) for all four models. The supervised probe (pink) rises sharply at 15--30\% depth and stays near ceiling thereafter - the familiarity information is present from early-middle layers on. The unsupervised dispersion metrics (IPR, spectral entropy) peak in a broad mid-depth band and oscillate layer-to-layer, showing that only some layers express the signal volumetrically. Dashed lines: per-model selection-aware floors (95th percentile of the permutation null with layer selection). Note that dispersion layers are MLP-hook indices ($0..N{-}1$) while probe layers are residual-stream indices ($0..N$); the relative-depth normalization is per layer family.}
\label{fig:depth}
\end{figure}

Figure~\ref{fig:depth} shows the per-layer picture on a shared relative-depth axis (layer index divided by the deepest layer of its family, which sidesteps the one-index offset between MLP and residual layer numbering). The probe saturates from roughly 20\% depth onward in every model: familiarity is linearly decodable through most of the network. Dispersion is choppier - individual layers alternate between strongly expressing and not expressing the signal - but its envelope forms a recognizable mid-depth band.

A band analysis (band = layers within 90\% of the per-model maximum AUROC) quantifies which metrics occupy a \emph{consistent} relative depth across all four models: spectral entropy shares a common band at relative depth $[0.48, 0.77]$, the probe at $[0.34, 1.00]$, logit-lens entropy at $[0.58, 0.68]$, and mean attention entropy overlaps only at the single depth $0.74$. IPR and effective rank have \emph{no} common band across the four models. The mid-to-late-depth ``knowledge band'' is therefore a real, scale-stable feature for entropy-flavored measures, but not a universal property of every dispersion statistic - consistent with the picture that different layers of different models express the underlying directional signal \citep{ferrando2025entity} to different volumetric degrees.

\subsection{Extended metrics: not all global measures are equal}
\label{sec:extended}

\begin{figure}[!tbp]
\centering
\includegraphics[width=0.8\textwidth]{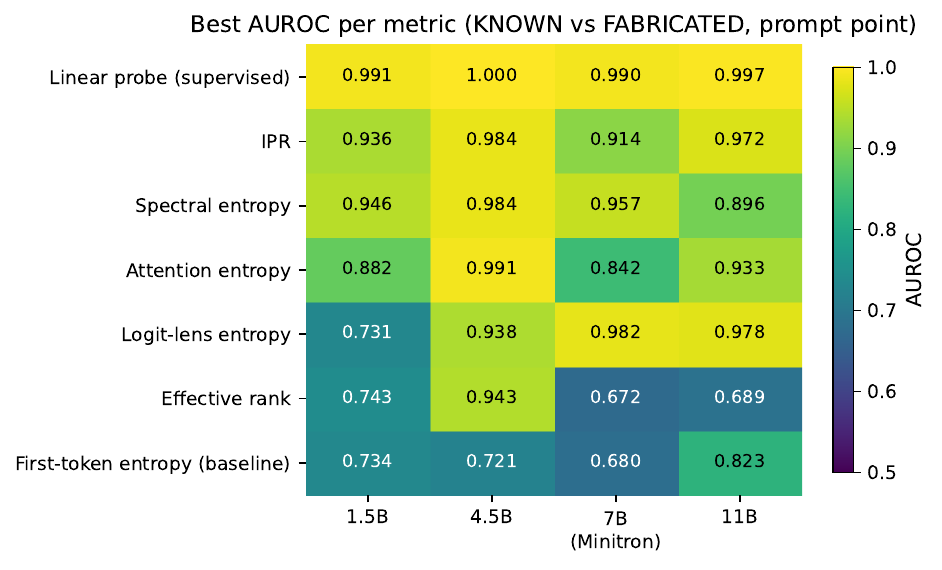}
\caption{Best per-metric AUROC (\known{} vs.\ \fabricated, prompt point) across the model sweep. MLP-side dispersion (IPR, spectral entropy) and the supervised probe are strong everywhere; attention entropy is robust across scale; logit-lens entropy strengthens with scale (0.72 at 1.5B to $\approx$0.98 at 7B/11B); effective rank is unstable (0.67--0.94). All internal metrics except effective rank at some scales clearly beat the output-side first-token-entropy baseline.}
\label{fig:metrics}
\end{figure}

\begin{table}[!tbp]
\centering
\footnotesize
\caption{Extended metrics: best AUROC per layer family (\known{} vs.\ \fabricated, prompt point), with best layer in parentheses. First-token entropy is an output-side baseline with no layer dimension.}
\label{tab:extended}
\begin{tabular}{lccccc}
\toprule
Model & Effective rank & Attn.\ entropy & Attn.\ entropy (norm.) & Logit-lens entropy & First-token ent. \\
\midrule
1.5B & 0.742 (24) & 0.879 (23) & 0.893 (23) & 0.722 (32) & 0.725 \\
4.5B & 0.942 (56) & 0.991 (43) & 0.923 (35) & 0.937 (35) & 0.718 \\
7B (Minitron) & 0.673 (27) & 0.839 (38) & 0.874 (38) & 0.983 (27) & 0.681 \\
11B & 0.692 (33) & 0.932 (24) & 0.925 (37) & 0.977 (19) & 0.830 \\
\bottomrule
\end{tabular}
\end{table}

Table~\ref{tab:extended} and Figure~\ref{fig:metrics} extend the comparison beyond MLP dispersion. The following three findings stand out. \textbf{Attention entropy is robust across scale} (0.839--0.991), consistent with the mechanistic account in which unknown entities suppress attribute-extraction attention, diffusing attention distributions \citep{ferrando2025entity,chuang2024lookback,binkowski2025lapeigvals}. \textbf{Logit-lens entropy grows with scale} (0.722 at 1.5B, 0.937 at 4.5B, then $\approx$0.98 at 7B and 11B): in larger models, how early the network commits to an answer becomes an increasingly reliable familiarity readout. \textbf{Effective rank is unstable} (0.673--0.942 with no scale trend and no common depth band): extending dispersion from a single activation vector to the token-span covariance does not survive as a reliable one-pass statistic in this setting, despite its kinship to multi-sample EigenScore \citep{chen2024inside}. Global measures are not interchangeable; the physically motivated single-vector measures and attention entropy transfer across scale, effective rank does not.

\subsection{The behavioral axis is much harder}
\label{sec:behavioral}

\begin{table}[!tbp]
\centering
\footnotesize
\caption{Behavioral axis, full grid (athletes): hallucination vs.\ correct \emph{within} the 42 \known{} entities, for every model and soft-judge threshold (prompt point; positives = hallucinating entities). Powered = both classes $\geq 8$. The floor column is the original \emph{single-permutation} floor, which understates the selection null (Section~\ref{sec:probe}); $\downarrow$ marks probe cells at or below even that floor. Dispersion best layers are shown because embedding-adjacent maxima (layers 0--1) flag likely noise.}
\label{tab:behavioral}
\setlength{\tabcolsep}{4.5pt}
\begin{tabular}{llcccccc}
\toprule
Model & Threshold & $n_\mathrm{hal}/n_\mathrm{cor}$ & IPR (layer) & Spectral ent.\ (layer) & Probe (layer) & Floor (1 perm) & Powered \\
\midrule
1.5B & 5/5 & 36/6 & 0.786 (6) & \textbf{0.829} (22) & 0.681 (22) $\downarrow$ & 0.806 & no \\
1.5B & $\geq$4/5 & 31/11 & \textbf{0.716} (27) & 0.684 (28) & 0.824 (5) & 0.666 & yes \\
1.5B & $\geq$3/5 & 23/19 & \textbf{0.698} (27) & 0.647 (28) & 0.584 (3) $\downarrow$ & 0.677 & yes \\
4.5B & 5/5 & 26/16 & \textbf{0.748} (0) & 0.740 (0) & 0.762 (50) & 0.675 & yes \\
4.5B & $\geq$4/5 & 17/25 & \textbf{0.735} (48) & 0.730 (48) & 0.807 (10) & 0.675 & yes \\
4.5B & $\geq$3/5 & 6/36 & \textbf{0.919} (5) & 0.910 (5) & 0.796 (50) & 0.708 & no \\
7B (Minitron) & 5/5 & 18/24 & \textbf{0.748} (33) & 0.745 (31) & 0.706 (2) $\downarrow$ & 0.819 & yes \\
7B (Minitron) & $\geq$4/5 & 12/30 & 0.700 (33) & \textbf{0.703} (33) & 0.897 (3) & 0.728 & yes \\
7B (Minitron) & $\geq$3/5 & 8/34 & \textbf{0.777} (38) & 0.748 (31) & 0.772 (21) $\downarrow$ & 0.787 & yes \\
11B & 5/5 & 9/33 & \textbf{0.681} (1) & 0.670 (1) & 0.929 (12) & 0.744 & yes \\
11B & $\geq$4/5 & 5/37 & \textbf{0.822} (9) & 0.794 (9) & 0.805 (33) $\downarrow$ & 0.870 & no \\
11B & $\geq$3/5 & 1/41 & \multicolumn{5}{c}{degenerate (single positive)} \\
\bottomrule
\end{tabular}
\end{table}

If dispersion measured factual reliability rather than mere familiarity, it should also separate correct answers from hallucinations \emph{within} the \known{} condition. Table~\ref{tab:behavioral} shows the full model $\times$ threshold grid, and its eight adequately powered cells (both classes $\geq 8$) are inconsistent in every direction that matters. At the Minitron-7B soft-5/5 cell, dispersion (0.748) nominally \emph{beats} the probe (0.706, which is itself below the single-permutation floor of 0.819); at the 4.5B soft-5/5 cell the ordering reverses (dispersion 0.748 vs.\ probe 0.762), and its dispersion maximum sits at layer 0. At the 1.5B $\geq$3/5 cell (the best-balanced cell in the grid, 23/19), the probe (0.584) falls \emph{below even its own single-permutation floor} (0.677): at 1.5B no reliable behavioral signal is readable at all. At the 11B soft-5/5 cell, the probe reaches 0.929 against a single-permutation floor of 0.744, while dispersion (0.681) is indistinguishable from the first-token-entropy baseline (0.701) and its best layer is layer 1 - embedding-adjacent, a flag that the number is noise rather than signal, which itself supports the negative reading. Given that selection-aware nulls reach 0.72--0.74 at $n=84$ on the detection contrast (Section~\ref{sec:probe}) and $n$ is only 42 here, we read the 11B probe cell as suggestive of \emph{some} weak, linearly decodable correctness signal, not as an established one; the remaining high values in the grid live in underpowered cells (minority class $\le 7$) and should not be read as signal.

We frame this as a mostly-negative result and, we believe, the paper's most useful distinction: \textbf{activation dispersion detects entity familiarity, not factual reliability}. Putting this into words, a one-pass dispersion gate can tell you the model has never heard of ``Roman Lewandowicz'', while it cannot tell you whether the model is about to attach a fabricated world championship to Robert Lewandowski.

\subsection{Semantic-entropy baseline: a complementary information source}
\label{sec:se}

\begin{table}[!tbp]
\centering
\footnotesize
\caption{Discrete semantic entropy (SE) over 5 samples with LLM clustering (zero clustering-fallback rows; athletes domain). AUROCs are separability, with 95\% bootstrap CIs where computed; the hallucination-label column uses judge labels over all 126 athlete entities (degenerate at 1.5B, where no entity is fully correct). Mean SE per condition shows self-consistency on \known{} rising with scale while \fabricated{}/\unknownreal{} stay inconsistent.}
\label{tab:se}
\setlength{\tabcolsep}{4pt}
\begin{tabular}{lcccccc}
\toprule
& \multicolumn{3}{c}{SE AUROC} & \multicolumn{3}{c}{Mean SE} \\
\cmidrule(lr){2-4}\cmidrule(lr){5-7}
Model & K vs.\ F & K vs.\ UR & halluc.\ (all 123) & \known & \unknownreal & \fabricated \\
\midrule
1.5B & 0.778 [0.678, 0.870] & 0.831 [0.743, 0.909] &  -  & 0.91 & 1.49 & 1.42 \\
4.5B & 0.759 [0.656, 0.851] & 0.833 [0.743, 0.909] & 0.776 & 0.59 & 1.25 & 1.12 \\
7B (Minitron) & 0.711 [0.599, 0.815] & 0.926 [0.867, 0.972] & 0.829 & 0.67 & 1.47 & 1.06 \\
11B & 0.832 [0.743, 0.911] & 0.932 [0.872, 0.978] & 0.872 & 0.26 & 1.29 & 0.97 \\
\bottomrule
\end{tabular}
\end{table}

\begin{figure}[!tbp]
\centering
\includegraphics[width=0.85\textwidth]{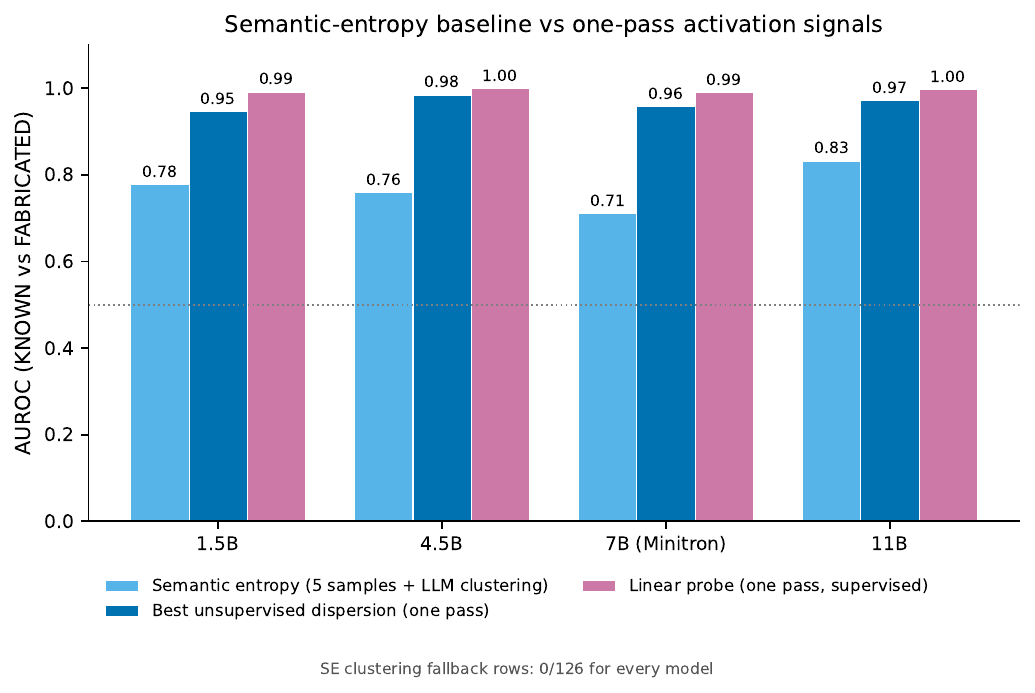}
\caption{Semantic entropy versus one-pass activation signals on \known{} vs.\ \fabricated. SE (5 samples + LLM clustering; zero fallback rows) reaches 0.71--0.83, in line with the literature range; single-pass unsupervised dispersion (0.94--0.98, non-overlapping 95\% CIs at every scale) and the supervised probe (0.99--1.00) exceed it on this familiarity contrast at a fifth of the sampling cost. On the correctness contrast the ordering reverses (Table~\ref{tab:se} and Section~\ref{sec:behavioral}): the two signals consume different information.}
\label{fig:se}
\end{figure}

Table~\ref{tab:se} and Figure~\ref{fig:se} compare against the standard multi-sample baseline. On the familiarity contrast, semantic entropy reaches 0.71--0.83 - squarely in the range reported for SE-style methods \citep{kuhn2023semantic,farquhar2024detecting} - while one-pass dispersion reaches 0.94--0.98; the 95\% CIs of the two do not overlap for any model (e.g., 11B: SE 0.832 [0.743, 0.911] vs.\ IPR 0.972 [0.939, 0.994]), and SE additionally costs five forward passes of the subject model plus clustering inference.

The comparison is, however, not apples-to-apples, and the two methods are better read as \emph{complementary information sources} than as competitors. Dispersion consumes the deterministic prefill and measures prompt-side familiarity. SE on the other hand, consumes the generation distribution (five samples at temperature 0.7) and measures answer-side consistency. On SE's home turf - the judge-labeled correctness contrast over all 126 athlete entities - SE reaches 0.78--0.87 (0.872 on 11B), precisely where dispersion fails (Section~\ref{sec:behavioral}). SE is also strong on \known{} vs.\ \unknownreal{} (0.83--0.93). And five samples is the low end for SE (ten or more are typical), so the SE numbers here should be read as a floor. Dispersion wins the familiarity contrast; SE wins the correctness contrast; a deployment stack may well want both.

The per-condition means expose a clean scaling effect: mean SE on \known{} falls from 0.91 (1.5B) through 0.59 and 0.67 to 0.26 (11B), while remaining high on \fabricated{} and \unknownreal{}. Larger models become increasingly \emph{self-consistent} about entities they know - even when the consistent answer still contains a fabricated detail, which is precisely why SE and factual correctness are not the same axis either.

\subsection{Generalization across entity types}
\label{sec:generalization}

\begin{figure}[!tbp]
\centering
\includegraphics[width=0.95\textwidth]{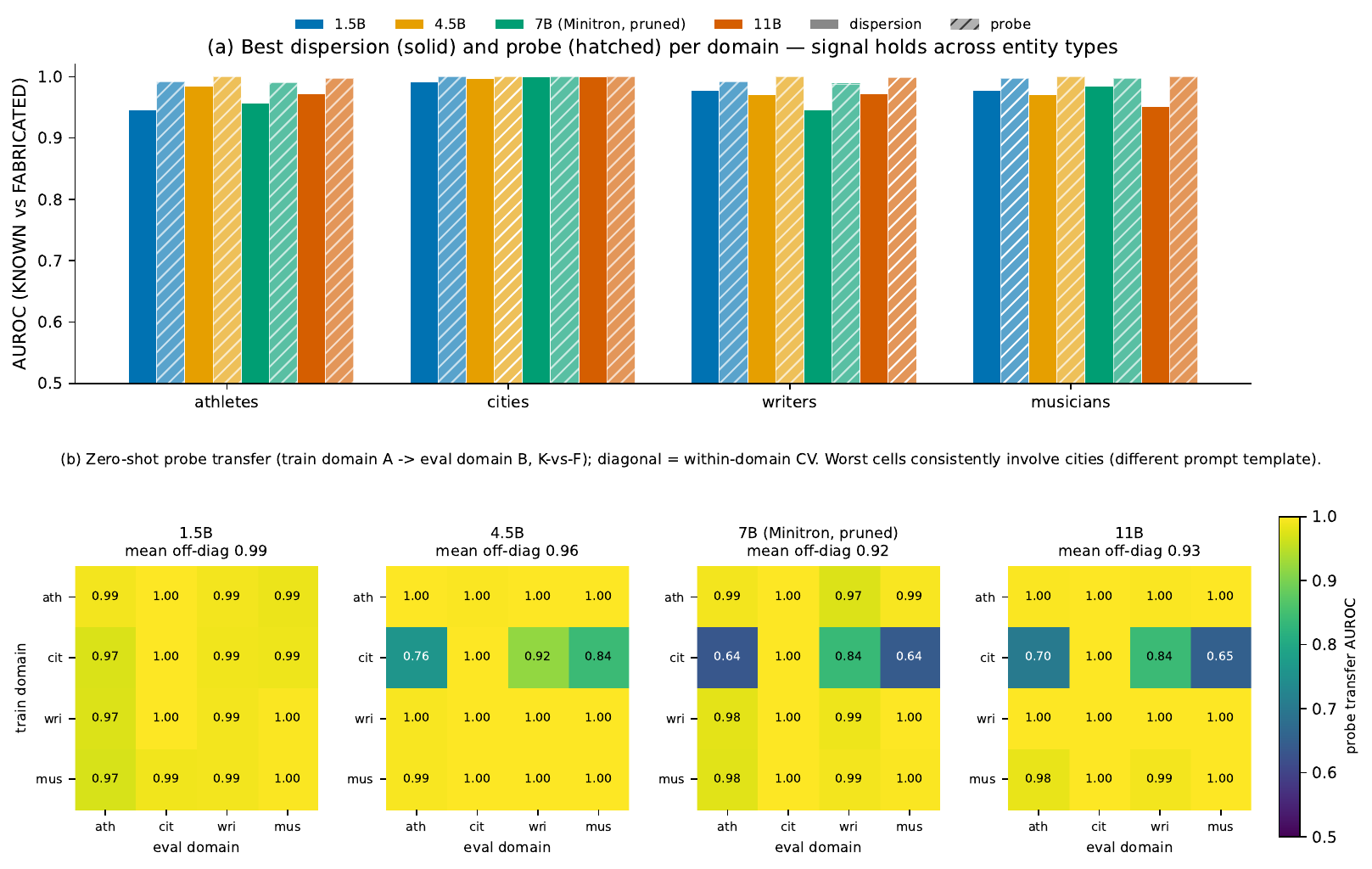}
\caption{The familiarity signal generalizes across entity types. \textbf{Left:} per-domain \known{}-vs-\fabricated{} separability (best dispersion / probe) for all four models - the one-pass detector works in every domain at every scale, with cities the easiest (probe reaches 1.000). \textbf{Right:} probe transfer matrices (train on domain A, evaluate on domain B, $4\times4$ per model). People-to-people transfer is near-perfect (0.985--0.998); the visible cold cells all involve cities, the one domain whose prompt template differs. A matched-template counterfactual shows the target-side drop is template-caused: under a shared neutral stem, writers$\rightarrow$cities transfer recovers to 0.999--1.000 at every scale (Section~\ref{sec:generalization}). The gap is $\approx$0 at 1.5B and grows to $+0.13$ at larger scales.}
\label{fig:domains}
\end{figure}

The results so far are all on athletes. To test whether the familiarity signal is a property of one entity type or of entity familiarity in general, we replicate the detection contrast on three further domains - cities, writers, and musicians - each with 42 \known{}/\unknownreal{}/\fabricated{} entities (condition-based labels only). Figure~\ref{fig:domains} summarizes; Table~\ref{tab:domains} gives the per-domain, per-model separability.

\begin{table}[!tbp]
\centering
\footnotesize
\caption{Cross-domain detection (\known{} vs.\ \fabricated, prompt point; $n=84$ per cell): best unsupervised dispersion metric / supervised probe AUROC, per model and domain. The one-pass signal separates known from fabricated entities in every domain at every scale; within-domain separability is if anything highest on cities (up to 1.000).}
\label{tab:domains}
\begin{tabular}{lcccc}
\toprule
Model & athletes & cities & writers & musicians \\
\midrule
1.5B & 0.946 / 0.991 & 0.991 / 1.000 & 0.978 / 0.991 & 0.978 / 0.997 \\
4.5B & 0.984 / 1.000 & 0.998 / 1.000 & 0.971 / 1.000 & 0.971 / 0.999 \\
7B (Minitron) & 0.957 / 0.990 & 1.000 / 1.000 & 0.946 / 0.988 & 0.984 / 0.997 \\
11B & 0.972 / 0.997 & 1.000 / 1.000 & 0.972 / 0.999 & 0.951 / 1.000 \\
\bottomrule
\end{tabular}
\end{table}

\paragraph{The signal transfers across domains.} Training the probe on one domain and evaluating on another (Figure~\ref{fig:domains}, right) preserves most of the separability: the mean off-diagonal transfer AUROC is 0.987 (1.5B), 0.959 (4.5B), 0.919 (7B), and 0.931 (11B). Transfer among the three \emph{people} domains (athletes, writers, musicians) is near-perfect at every scale (0.985--0.998). This parallels the cross-entity-type generalization reported by \citet{ferrando2025entity} for their known-entity direction, while extending it to a global, unsupervised statistic on a different model family and language - and it is the opposite of the cross-dataset generalization failures reported for supervised truthfulness probes \citep{orgad2025llmsknow}, consistent with familiarity being a more portable representation-level quantity than answer truthfulness.

\paragraph{The transfer drops are template-caused: a matched-template counterfactual.} Every large off-diagonal drop involves the cities domain, which is the only domain using the \emph{Czym jest} (``what is'') stem rather than \emph{Kim jest} (``who is''); the worst cells are cities$\rightarrow$athletes at 0.637 (Minitron) and cities$\rightarrow$musicians at 0.652 (11B), and the people-to-people versus cities-involving gap is $+0.078$ (4.5B), $+0.133$ (Minitron), and $+0.129$ (11B), but only $-0.005$ at 1.5B. Template and entity type are confounded for cities in the main matrix, so we ran a counterfactual control: cities and writers re-extracted under the shared neutral stem \emph{Co wiesz o \{entity\}?} (Section~\ref{sec:dataset}), which matches the template while varying entity type, on all four models. The neutral extractions are internally valid (within-domain K-vs-F probe 0.99--1.00, dispersion 0.96--1.00), and the decisive entity-type-varying cell recovers to ceiling: writers$\rightarrow$cities under the matched stem reaches probe transfer 0.999/1.000/1.000/1.000 (1.5B/4.5B/7B/11B), indistinguishable from the people$\rightarrow$people reference (0.98--1.00). The target-side drop into cities is therefore \emph{template-caused, not an entity-type effect}. One asymmetry survives the matched template: cities \emph{as source} transfers outward less well (cities$\rightarrow$writers probe 0.978/0.918/0.825/0.816), consistent with a tighter, lower-variance places manifold whose fitted probes generalize outward less well - not with places being harder to represent (within-cities separability remains the highest of any domain, Table~\ref{tab:domains}) - and matching the original worst cells, which all have cities as the \emph{training} domain. The template caveat is one \citet{ferrando2025entity} did not encounter, since their cross-entity-type transfer held the prompt frame fixed.

\paragraph{The knowledge band sits at a common relative depth across domains.} A layer-band analysis (band = relative depths within 90\% of the per-domain maximum AUROC) shows the detector occupies a shared mid-depth band across all four domains: for the probe, the cross-domain intersection is relative depth $[0.16, 1.00]$--$[0.38, 1.00]$ with overlap/union rising from 0.74 (1.5B) to 0.91 (11B); for dispersion the common band is roughly $[0.4, 0.85]$, loosest on the 11B dispersion metric (overlap/union 0.28, best-depth spread 0.45). The band is thus a scale-stable, largely domain-independent feature for the probe and for entropy-flavored dispersion, and noisiest exactly where the individual dispersion metric was already least stable (11B, Section~\ref{sec:scaling}).

\begin{figure}[!tbp]
\centering
\includegraphics[width=0.9\textwidth]{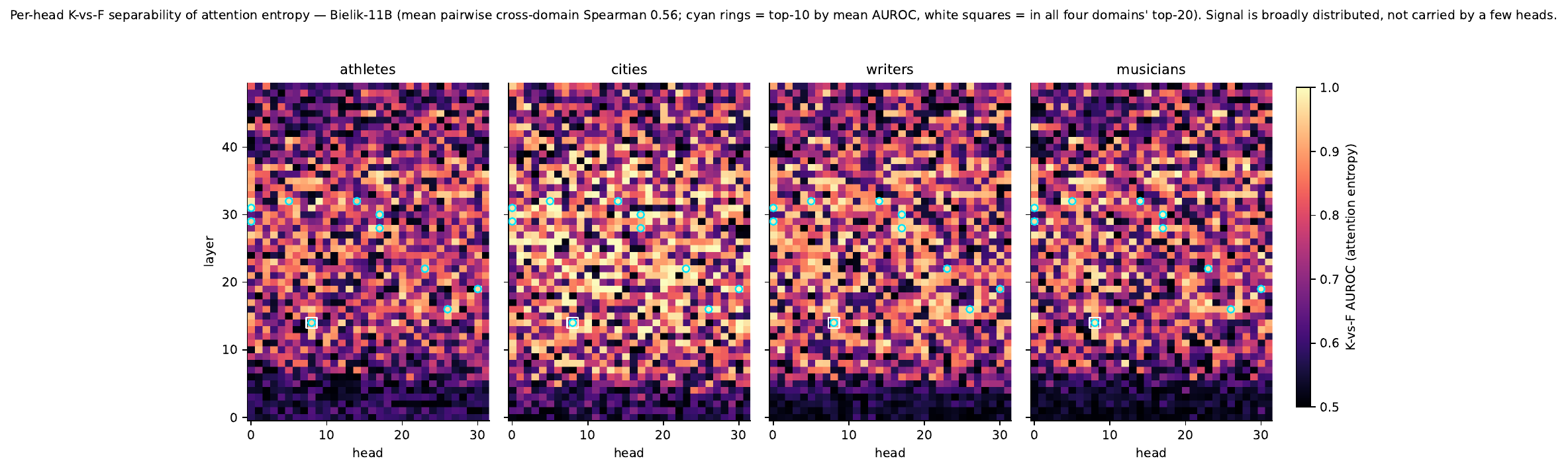}
\caption{Per-head attention-entropy separability (\known{} vs.\ \fabricated, prompt point) is diffuse. For each model, heads are ranked by their K-vs-F AUROC within a domain, and cross-domain consistency is measured by the Spearman correlation of the full per-head AUROC vectors and by top-20 head-set overlap. Only 1--2 heads appear in all four domains' top-20; mean pairwise Spearman rises with scale (0.27$\rightarrow$0.56) but stays modest. The signal is carried by many partially overlapping heads at relative depth $\approx$0.6--0.8, not by a small fixed set.}
\label{fig:heads}
\end{figure}

\paragraph{The signal is diffuse across attention heads.} Ranking every head by the \known{}-vs-\fabricated{} separability of its prompt-point attention entropy (Figure~\ref{fig:heads}), only 1--2 heads sit in all four domains' top-20 at any scale, and the mean pairwise cross-domain Spearman correlation of the per-head AUROC vectors rises only modestly with scale: 0.27 (1.5B, 384 heads), 0.50 (4.5B, 960), 0.52 (Minitron, 1280), 0.56 (11B, 1600). The best cross-domain heads sit at relative depth 0.63--0.80 (e.g., 11B L31H0, mean 0.986). The attention-entropy familiarity signal is therefore largely \emph{diffuse} - spread across many heads that overlap only partially between domains - with the overlap strengthening but staying modest as models grow. This does not contradict the concentrated circuits reported in causal work - the attribute-extraction heads modulated by entity-recognition directions \citep{ferrando2025entity}, or the sparse, causally necessary retrieval heads of \citet{wu2025retrievalheads}: those claims are mediational, ours is marginal correlational separability, and a concentrated causal circuit is fully compatible with diffuse marginal separability once its effects propagate downstream. What the diffuseness does establish is that our global statistics read a broadly distributed correlate of familiarity, not the circuit itself. As on athletes, effective rank remains the weakest metric across domains.

\section{Discussion}
\label{sec:discussion}

\paragraph{Two phenomena on different scaling curves.}
The central empirical claim is a dissociation. Entity familiarity at the representation level is already at ceiling on our contrast at 1.5B: cheap, label-free dispersion statistics read it out at 0.94+, and a linear probe at $\approx$0.99 (whether familiarity as a graded quantity still improves with scale is not measurable on a famous-vs-fabricated contrast that every model aces). Behavioral factual reliability - answering about familiar entities without invented decorations - is nowhere near solved at 1.5B and improves steeply through 11B. These curves cannot be two views of one underlying quantity; a detector of the first is not a detector of the second, as Section~\ref{sec:behavioral} shows directly. This refines the localization intuition that motivated the study: the delocalized-activation signature marks \emph{missing entities}, not \emph{missing facts}. It also gives a coarse-grained, purely volumetric confirmation of the mechanistic picture in which entity-recognition circuitry \citep{ferrando2025entity,lindsey2025biology} is present and functional even in small models, while the knowledge those circuits gate remains scale-limited - consistent with frequency-based accounts of factual recall, in which long-tail facts arrive only with capacity and data repetition \citep{mallen2023trust,kandpal2023longtail} while entity recognition arrives much earlier. The familiarity axis is moreover not an athletes-specific curiosity: it reproduces across cities, writers, and musicians and transfers between them (Section~\ref{sec:generalization}), so the detector reads a general property of entity familiarity rather than a domain-specific lexical quirk. Though, being diffuse across heads, it is a coarse volumetric shadow of the circuitry, not a localization of it. Nor is it a Bielik-specific quirk. A compact control on a non-Polish-centric family (Gemma-4 at two sizes, identical dataset and pipeline; Appendix~\ref{app:gemma}) replicates both the familiarity signal and the dissociation qualitatively: weaker, and still rising with scale, weakest exactly in the domains whose famous tier is most Polish-local, which is what a signal that tracks \emph{actual} model familiarity should do.

\paragraph{The near-zero-refusal regime.}
An answer-level LLM audit (Claude Haiku 4.5) of all 2{,}520 sampled athlete answers - including 42 entities that do not exist and 42 the models demonstrably know nothing about - finds exactly 2 refusals and 1 hedged answer: 99.88\% of answers are direct assertions. The models fluently biografize fabricated athletes while their own activations carry a near-perfectly decodable ``I don't know this entity'' signal. Strikingly, both refusals come from the 11B model (one \unknownreal{}, one \fabricated{} entity; the single hedge is from Minitron-7B): \emph{only the largest model ever declines}, and even it at a per-mille rate. This sharpens rather than weakens the abstention-gap point - the capacity to refuse emerges with scale, exactly as the steering results of \citet{ferrando2025entity} (where activating the unknown-entity direction \emph{causes} refusal) and the default-refusal circuits of \citet{lindsey2025biology} would lead one to expect, yet it remains behaviorally negligible at 11B while the internal familiarity signal is already near-ceiling at 1.5B. One qualifier is owed: the prompt commands a one-sentence answer (\emph{``Odpowiedz jednym zdaniem.''}), which plausibly pushes an instruction-following model away from abstaining, so the regime is partly a property of the prompt (Section~\ref{sec:limitations}). The Gemma-4 control bounds that qualifier: under the identical prompts, Gemma-4-12B abstains \emph{condition-selectively} - 24/42 \fabricated{} and 28/42 \unknownreal{} athletes draw all-five-sample refusals, versus 7/42 \known{} (Appendix~\ref{app:gemma}) - so the prompt does not preclude abstention, and a model of this scale range \emph{can} surface the familiarity representation behaviorally. That Bielik does not is therefore a \textbf{tuning/alignment property}, not an inevitability of the representational signal or the scale range. The representational precondition for calibrated abstention exists at every scale; the behavior barely exists at any Bielik scale. \textbf{That is an actionable alignment gap for Bielik family rather than a capability gap}.

\paragraph{Practical implication: one-pass familiarity gating.}
For Polish LLM deployments, the immediate application is a cheap gate. Compute IPR/spectral entropy at the prompt's last token during the prefill pass (essentially free), and route high-dispersion queries to retrieval, abstention, or escalation. Unlike SelfCheckGPT or semantic entropy \citep{manakul2023selfcheckgpt,farquhar2024detecting}, this requires no extra samples and no auxiliary model, and unlike probe-based detectors \citep{azaria2023internal,kossen2024semantic} it needs no labeled training data - though where labels are available, the probe's near-ceiling performance argues for using both. Two caveats apply. The gate reads familiarity, not existence: it scores real-but-obscure and fabricated entities similarly (\unknownreal{} vs.\ \fabricated{} 0.65--0.82 across domains), which is acceptable for routing - both belong in retrieval or abstention - but rules out using it to tell ``obscure but real'' from ``nonexistent.'' And, per Section~\ref{sec:behavioral}, the gate flags unfamiliar \emph{entities}; it must not be sold as a hallucination detector for content about familiar ones.

\section{Limitations}
\label{sec:limitations}

\textbf{Lexical confound (C1), quantified.} Fabricated names are rare strings, and rare tokens alone perturb activation statistics. The length matching between \fabricated{} and \known{} is only approximate: for athletes, token count alone reaches K-vs-F AUROC 0.681/0.612 (small/large tokenizer), and the \unknownreal{}-vs-\fabricated{} dispersion separation of 0.70--0.82 upper-bounds the lexical-plus-plausibility component (Section~\ref{sec:controls}). The \known{}-vs-\unknownreal{} control (0.97--1.00), free of fabricated strings, shows familiarity dominates; residual lexical contributions to the headline cannot be excluded, but they cannot be the story. The three new domains are matched at least as tightly (token-count K-vs-F AUROC 0.49--0.64; Section~\ref{sec:dataset}), so the generalization result is not carried by a lexical artifact either. \textbf{Sample size.} Each detection contrast has $n=84$ (42 per class; 42 per class on the behavioral axis); bootstrap CIs on headline cells span roughly $\pm$0.03--0.06 and selection-aware permutation nulls reach 0.70--0.74, so single-model AUROC differences below $\approx$0.05 should not be interpreted. \textbf{Behavioral, SE, and refusal claims are athletes-only.} The behavioral axis, semantic-entropy comparison, and refusal audit are measured on athletes; the cities, writers, and musicians domains carry condition-based detection labels only, so the dissociation and abstention-gap claims are demonstrated on one domain even though the detection signal itself is shown to generalize. \textbf{Template effects in cross-domain transfer.} The cities domain uses a different prompt stem (\emph{Czym jest} rather than \emph{Kim jest}), confounding template and entity type in the main transfer matrix. The matched-template control (Section~\ref{sec:generalization}) resolves the target-side attribution causally - transfer into cities recovers to ceiling under a shared neutral stem - but two caveats remain: the control re-extracted only cities and writers under a single alternative stem, and the residual cities-as-source asymmetry (probe 0.82--0.98 under matched templates, widening with scale) is characterized, not explained; we read it as a manifold-tightness effect, which a targeted geometry analysis could test directly. \textbf{Language and model family.} The main study is one Polish family. A compact control on a non-Polish-centric family (Gemma-4 at two sizes, same Polish dataset, pipeline, and judges; Appendix~\ref{app:gemma}) qualitatively replicates the familiarity signal and the dissociation, which removes the sharpest version of this concern - but it is two checkpoints of one additional family on one language; the full multi-family, multi-language treatment (including an English translation of the dataset) is future work (already in progress). \textbf{Per-head diffuseness.} The attention-entropy signal is spread across many partially overlapping heads (only 1--2 shared across all four domains' top-20; cross-domain Spearman 0.27--0.56), which qualifies any mechanistic reading: our global statistics are a coarse volumetric shadow of the entity-recognition circuitry \citep{ferrando2025entity}, not a localization of a small head set. \textbf{Fabricated non-existence.} The 168 invented names (42 per domain) were screened. Automated pool-stage filtering, manual morphology/confusability/notability checks, and, for the three new domains, per-name search-engine verification with a logged replacement trail - but not exhaustively verified against all real persons and places; rare collisions with obscure real individuals are possible, though the models would, with high likelihood, not know such individuals either, so labels are robust to this failure mode. \textbf{Judge stack and circularity.} One model family occupies every rung of the evaluation ladder: strict and soft judging (Claude Opus 4.8), semantic-entropy clustering (Claude Sonnet 5), and the refusal audit (Claude Haiku 4.5). A family-level blind spot about Polish sports biography would contaminate the behavioral counts, the SE correctness column, and the refusal counts simultaneously; no non-Claude judge was run (but is planned for future work). This exposure extends to the behavioral \emph{scaling} claim itself: the 0$\rightarrow$2$\rightarrow$10$\rightarrow$19 strict and 6$\rightarrow$16$\rightarrow$24$\rightarrow$33 soft curves rest entirely on one judge family: the monotone trend is corroborated by two independent rubrics and is too large to plausibly be judge noise, but it has not been confirmed by a second judge family. The strict judge is additionally a bare one-line prompt whose parse failures default to \emph{incorrect} (Section~\ref{sec:labeling}), so strict counts conflate confabulation with truncation and parsing artifacts - the soft rubric mitigates but does not eliminate this. Manual inspection of 615 soft-judge verdicts (the \known{} answers of the three dense models) surfaced 2 anomalous cases ($\approx$0.3\%): one flag of a 4.5B answer crediting Ewa Swoboda with a 2018 European indoor title, where the flag was arguably \emph{correct} (no European championships were held in 2018) and the strict judge's pass was the error; and one false flag of an 11B answer about Agnieszka Radwańska that was truncated mid-sentence by the 64-token cap. We report this as a manual-inspection finding, not an on-disk artifact. \textbf{Refusal operationalization and prompt pressure.} Refusal counts are answer-level LLM classifications, not human annotations; and the one-sentence instruction in the prompt plausibly discourages abstention, so the near-zero refusal rate is partly a property of the prompt. The Gemma-4 control bounds this concern - under the identical one-sentence prompt, Gemma-4-12B refuses condition-selectively (Appendix~\ref{app:gemma}), so the prompt does not \emph{preclude} abstention - but a prompt-variant ablation on Bielik itself (without the instruction, or with an explicit ``say so if you are unsure'') was still not run. \textbf{Winsorization sensitivity.} The dispersion quantile $q=0.99$ is the most favorable of the four settings swept, the sweep covered only the 1.5B model on the round-1 athletes list (41 per condition), and per-layer values shift substantially with $q$ (Section~\ref{sec:metrics}); the familiarity signal itself persists at every setting. \textbf{Underpowered behavioral cells.} Four of the twelve grid cells have a minority class below 8 (1.5B 5/5; 4.5B $\geq$3/5; 11B $\geq$4/5 and $\geq$3/5); conclusions rest on the eight powered cells, which are themselves mutually inconsistent (Section~\ref{sec:behavioral}) - the honest summary is that no reliable behavioral signal has been demonstrated, not that one was ruled out. \textbf{Minitron.} The 7B point is a pruned/distilled variant \citep{muralidharan2024minitron}, not a dense family member; it tracks the dense scale curve on strict and soft-5/5 criteria but is non-monotonic at soft $\geq$3/5 (34 vs.\ 36 for 4.5B), and its layer count and depth profile are not directly comparable.

\section{Reproducibility}
\label{sec:repro}

The full pipeline is included in the aforementioned public code repository: dataset construction, answer sampling and judging, activation extraction, and analysis, is a single command per model. Experiments ran on one Apple-silicon workstation (MPS backend, bf16); no GPU cluster is required at these scales. With Python tooling via \texttt{uv}:

\begin{quote}
\small
\begin{tabular}{l}
\texttt{uv sync --extra dev} \\
\texttt{uv run pytest -q}\hfill \emph{(57 unit tests)} \\
\texttt{uv run python run\_mvp.py}\hfill \emph{(defaults to 1.5B)} \\
\texttt{BIELIK\_MODEL\_ID=speakleash/Bielik-11B-v3.0-Instruct uv run python run\_mvp.py}
\end{tabular}
\end{quote}

\begin{sloppypar}
Hugging Face access to the gated Bielik v3.0 checkpoints and an Anthropic API key (judging, semantic-entropy clustering) are configured via \texttt{environment variables}. Per-model artifacts are written to \texttt{results/<model>/} (\texttt{signals.parquet}, \texttt{hidden\_states.npz}, \texttt{extended\_signals.parquet}, \texttt{semantic\_entropy.parquet}, per-layer AUROC figures) and \texttt{data/<model>/} (\texttt{labeled.parquet} with all five sampled answers per entity, plus soft-judge labels); the three additional domains live under \texttt{data/<model>/domains/<domain>/} and \texttt{results/<model>/domains/<domain>/} with condition-based signals only. Judged with Claude Opus 4.8 (strict and soft rubrics); semantic-entropy clustering with Claude Sonnet 5; refusal audit with Claude Haiku 4.5. One reproducibility asymmetry should be stated plainly: answer sampling was unseeded (temperature 0.7), so the behavioral counts, SE values, and refusal audit are properties of the frozen answer artifacts shipped with the repository rather than of a fresh run; the activation/dispersion side is deterministic and exactly re-derivable. All numbers in this paper are recomputed from the raw artifacts by included consolidation scripts.
\end{sloppypar}

\section{Conclusion}
\label{sec:conclusion}

Bielik knows what it doesn't know. At every scale we tested, and legibly enough that two closed-form, unsupervised dispersion statistics read the distinction from a single forward pass at 0.94--0.98 AUROC. What Bielik does \emph{not} know, below 11B, is the facts themselves: behavioral reliability about familiar athletes climbs from 0/42 to 19/42 across the same sweep, and no volumetric statistic we tested tracks it. Between these two curves sits the deployment problem: a model family that says ``I don't know'' twice in 2{,}520 answers while its activations say exactly that, almost perfectly, at every scale and in every entity domain we tested. Closing that gap - by steering, fine-tuning, or simple prefill-time gating - looks less like a research puzzle and more like unfinished engineering.

\clearpage
\bibliographystyle{plainnat}
{\small
\bibliography{refs}

\begin{thebibliography}{38}
\providecommand{\natexlab}[1]{#1}
\providecommand{\url}[1]{\texttt{#1}}
\expandafter\ifx\csname urlstyle\endcsname\relax
  \providecommand{\doi}[1]{doi: #1}\else
  \providecommand{\doi}{doi: \begingroup \urlstyle{rm}\Url}\fi

\bibitem[Abdaljalil et~al.(2025)Abdaljalil, Kurban, and
  Serpedin]{abdaljalil2025halluverse}
Samir Abdaljalil, Hasan Kurban, and Erchin Serpedin.
\newblock {HalluVerse25}: Fine-grained multilingual benchmark dataset for {LLM}
  hallucinations, 2025.
\newblock URL \url{https://arxiv.org/abs/2503.07833}.

\bibitem[Adams(1979)]{adams1979hitchhiker}
Douglas Adams.
\newblock \emph{The Hitchhiker's Guide to the Galaxy}.
\newblock Pan Books, London, 1979.

\bibitem[Alvi et~al.(2026)Alvi, Sayeedi, and Sayeedi]{alvi2026multihaludet}
Riasad Alvi, Nurul~Labib Sayeedi, and Md. Faiyaz~Abdullah Sayeedi.
\newblock {MultiHaluDet}: Multilingual hallucination detection via {LLM} hidden
  state probing.
\newblock In \emph{Proceedings of the 1st Workshop on Multilinguality in the
  Era of Large Language Models (MeLLM 2026)}. Association for Computational
  Linguistics, 2026.
\newblock URL \url{https://aclanthology.org/2026.mellm-1.6/}.

\bibitem[Anderson(1958)]{anderson1958absence}
P.~W. Anderson.
\newblock Absence of diffusion in certain random lattices.
\newblock \emph{Physical Review}, 109\penalty0 (5):\penalty0 1492--1505, 1958.
\newblock \doi{10.1103/PhysRev.109.1492}.

\bibitem[Azaria and Mitchell(2023)]{azaria2023internal}
Amos Azaria and Tom Mitchell.
\newblock The internal state of an {LLM} knows when it's lying.
\newblock In \emph{Findings of the Association for Computational Linguistics:
  EMNLP 2023}, pages 967--976. Association for Computational Linguistics, 2023.
\newblock \doi{10.18653/v1/2023.findings-emnlp.68}.
\newblock URL \url{https://aclanthology.org/2023.findings-emnlp.68/}.

\bibitem[Binkowski et~al.(2025)Binkowski, Janiak, Sawczyn, Gabrys, and
  Kajdanowicz]{binkowski2025lapeigvals}
Jakub Binkowski, Denis Janiak, Albert Sawczyn, Bogdan Gabrys, and Tomasz
  Kajdanowicz.
\newblock Hallucination detection in {LLM}s using spectral features of
  attention maps.
\newblock In \emph{Proceedings of the 2025 Conference on Empirical Methods in
  Natural Language Processing (EMNLP)}, 2025.
\newblock URL \url{https://arxiv.org/abs/2502.17598}.

\bibitem[Chen et~al.(2024)Chen, Liu, Chen, Gu, Wu, Tao, Fu, and
  Ye]{chen2024inside}
Chao Chen, Kai Liu, Ze~Chen, Yi~Gu, Yue Wu, Mingyuan Tao, Zhihang Fu, and
  Jieping Ye.
\newblock {INSIDE}: {LLM}'s internal states retain the power of hallucination
  detection.
\newblock In \emph{International Conference on Learning Representations
  (ICLR)}, 2024.
\newblock URL \url{https://openreview.net/forum?id=Zj12nzlQbz}.

\bibitem[Chuang et~al.(2024)Chuang, Qiu, Hsieh, Krishna, Kim, and
  Glass]{chuang2024lookback}
Yung-Sung Chuang, Linlu Qiu, Cheng-Yu Hsieh, Ranjay Krishna, Yoon Kim, and
  James~R. Glass.
\newblock Lookback lens: Detecting and mitigating contextual hallucinations in
  large language models using only attention maps.
\newblock In \emph{Proceedings of the 2024 Conference on Empirical Methods in
  Natural Language Processing (EMNLP)}, pages 1419--1436. Association for
  Computational Linguistics, 2024.
\newblock URL \url{https://aclanthology.org/2024.emnlp-main.84/}.

\bibitem[Ding et~al.(2025)Ding, Zhu, Xia, Wu, Chen, Liu, and
  Wang]{ding2025d2hscore}
Yue Ding, Xiaofang Zhu, Tianze Xia, Junfei Wu, Xinlong Chen, Qiang Liu, and
  Liang Wang.
\newblock {D\textsuperscript{2}HScore}: Reasoning-aware hallucination detection
  via semantic breadth and depth analysis in {LLM}s, 2025.
\newblock URL \url{https://arxiv.org/abs/2509.11569}.
\newblock Under review.

\bibitem[Du et~al.(2024)Du, Xiao, and Li]{du2024haloscope}
Xuefeng Du, Chaowei Xiao, and Yixuan Li.
\newblock {HaloScope}: Harnessing unlabeled {LLM} generations for hallucination
  detection.
\newblock In \emph{Advances in Neural Information Processing Systems
  (NeurIPS)}, 2024.
\newblock URL \url{https://openreview.net/forum?id=nfK0ZXFFSn}.

\bibitem[Ettori et~al.(2025)Ettori, Darabi, Tayebati, Krishnan, Subedar,
  Tickoo, and Trivedi]{ettori2025eigentrack}
Davide Ettori, Nastaran Darabi, Sina Tayebati, Ranganath Krishnan, Mahesh
  Subedar, Omesh Tickoo, and Amit~Ranjan Trivedi.
\newblock {EigenTrack}: Spectral activation feature tracking for hallucination
  and out-of-distribution detection in {LLM}s and {VLM}s, 2025.
\newblock URL \url{https://arxiv.org/abs/2509.15735}.
\newblock Submitted to ICASSP 2026.

\bibitem[Farquhar et~al.(2024)Farquhar, Kossen, Kuhn, and
  Gal]{farquhar2024detecting}
Sebastian Farquhar, Jannik Kossen, Lorenz Kuhn, and Yarin Gal.
\newblock Detecting hallucinations in large language models using semantic
  entropy.
\newblock \emph{Nature}, 630\penalty0 (8017):\penalty0 625--630, 2024.
\newblock \doi{10.1038/s41586-024-07421-0}.

\bibitem[Ferrando et~al.(2025)Ferrando, Obeso, Rajamanoharan, and
  Nanda]{ferrando2025entity}
Javier Ferrando, Oscar Obeso, Senthooran Rajamanoharan, and Neel Nanda.
\newblock Do {I} know this entity? {K}nowledge awareness and hallucinations in
  language models.
\newblock In \emph{International Conference on Learning Representations
  (ICLR)}, 2025.
\newblock URL \url{https://arxiv.org/abs/2411.14257}.

\bibitem[Kadavath et~al.(2022)Kadavath, Conerly, Askell, Henighan, Drain,
  Perez, Schiefer, Hatfield-Dodds, DasSarma, Tran-Johnson, Johnston, El-Showk,
  Jones, Elhage, Hume, Chen, Bai, Bowman, Fort, Ganguli, Hernandez, Jacobson,
  Kernion, Kravec, Lovitt, Ndousse, Olsson, Ringer, Amodei, Brown, Clark,
  Joseph, Mann, McCandlish, Olah, and Kaplan]{kadavath2022languagemodels}
Saurav Kadavath, Tom Conerly, Amanda Askell, Tom Henighan, Dawn Drain, Ethan
  Perez, Nicholas Schiefer, Zac Hatfield-Dodds, Nova DasSarma, Eli
  Tran-Johnson, Scott Johnston, Sheer El-Showk, Andy Jones, Nelson Elhage,
  Tristan Hume, Anna Chen, Yuntao Bai, Sam Bowman, Stanislav Fort, Deep
  Ganguli, Danny Hernandez, Josh Jacobson, Jackson Kernion, Shauna Kravec,
  Liane Lovitt, Kamal Ndousse, Catherine Olsson, Sam Ringer, Dario Amodei, Tom
  Brown, Jack Clark, Nicholas Joseph, Ben Mann, Sam McCandlish, Chris Olah, and
  Jared Kaplan.
\newblock Language models (mostly) know what they know, 2022.
\newblock URL \url{https://arxiv.org/abs/2207.05221}.

\bibitem[Kandpal et~al.(2023)Kandpal, Deng, Roberts, Wallace, and
  Raffel]{kandpal2023longtail}
Nikhil Kandpal, Haikang Deng, Adam Roberts, Eric Wallace, and Colin Raffel.
\newblock Large language models struggle to learn long-tail knowledge.
\newblock In \emph{Proceedings of the 40th International Conference on Machine
  Learning (ICML)}, volume 202 of \emph{PMLR}, pages 15696--15707, 2023.
\newblock URL \url{https://proceedings.mlr.press/v202/kandpal23a.html}.

\bibitem[Kinas et~al.(2026)Kinas, Kiszczak, Perez, Ociepa, Flis, Wr{\'o}bel,
  and Gwo{\'z}dziej]{kinas2026minitron}
Remigiusz Kinas, Pawe{\l} Kiszczak, Sergio~P. Perez, Krzysztof Ociepa,
  {\L}ukasz Flis, Krzysztof Wr{\'o}bel, and Adrian Gwo{\'z}dziej.
\newblock Bielik-minitron-7b: Compressing large language models via structured
  pruning and knowledge distillation for the polish language, 2026.
\newblock arXiv:2603.11881.

\bibitem[Kossen et~al.(2024)Kossen, Han, Razzak, Schut, Malik, and
  Gal]{kossen2024semantic}
Jannik Kossen, Jiatong Han, Muhammed Razzak, Lisa Schut, Shreshth Malik, and
  Yarin Gal.
\newblock Semantic entropy probes: Robust and cheap hallucination detection in
  {LLM}s, 2024.
\newblock URL \url{https://arxiv.org/abs/2406.15927}.
\newblock ICML 2024 Workshop on Foundation Models in the Wild.

\bibitem[Kuhn et~al.(2023)Kuhn, Gal, and Farquhar]{kuhn2023semantic}
Lorenz Kuhn, Yarin Gal, and Sebastian Farquhar.
\newblock Semantic uncertainty: Linguistic invariances for uncertainty
  estimation in natural language generation.
\newblock In \emph{International Conference on Learning Representations
  (ICLR)}, 2023.
\newblock URL \url{https://arxiv.org/abs/2302.09664}.

\bibitem[Li et~al.(2023)Li, You, Bhojanapalli, Li, Rawat, Reddi, Ye, Chern, Yu,
  Guo, and Kumar]{li2023lazy}
Zonglin Li, Chong You, Srinadh Bhojanapalli, Daliang Li, Ankit~Singh Rawat,
  Sashank~J. Reddi, Ke~Ye, Felix Chern, Felix Yu, Ruiqi Guo, and Sanjiv Kumar.
\newblock The lazy neuron phenomenon: On emergence of activation sparsity in
  transformers.
\newblock In \emph{International Conference on Learning Representations
  (ICLR)}, 2023.
\newblock URL \url{https://openreview.net/forum?id=TJ2nxciYCk-}.

\bibitem[Lindsey et~al.(2025)Lindsey, Gurnee, Ameisen, Chen, Pearce, Turner,
  Citro, Abrahams, Carter, Hosmer, Marcus, Sklar, Templeton, Bricken,
  McDougall, Cunningham, Henighan, Jermyn, Jones, Persic, Qi, Thompson,
  Zimmerman, Rivoire, Conerly, Olah, and Batson]{lindsey2025biology}
Jack Lindsey, Wes Gurnee, Emmanuel Ameisen, Brian Chen, Adam Pearce,
  Nicholas~L. Turner, Craig Citro, David Abrahams, Shan Carter, Basil Hosmer,
  Jonathan Marcus, Michael Sklar, Adly Templeton, Trenton Bricken, Callum
  McDougall, Hoagy Cunningham, Thomas Henighan, Adam Jermyn, Andy Jones, Andrew
  Persic, Zhenyi Qi, T.~Ben Thompson, Sam Zimmerman, Kelley Rivoire, Thomas
  Conerly, Chris Olah, and Joshua Batson.
\newblock On the biology of a large language model.
\newblock Transformer Circuits Thread, 2025.
\newblock URL
  \url{https://transformer-circuits.pub/2025/attribution-graphs/biology.html}.
\newblock Published March 27, 2025.

\bibitem[Liu et~al.(2023)Liu, Wang, Dao, Zhou, Yuan, Song, Shrivastava, Zhang,
  Tian, R{\'e}, and Chen]{liu2023dejavu}
Zichang Liu, Jue Wang, Tri Dao, Tianyi Zhou, Binhang Yuan, Zhao Song, Anshumali
  Shrivastava, Ce~Zhang, Yuandong Tian, Christopher R{\'e}, and Beidi Chen.
\newblock Deja vu: Contextual sparsity for efficient {LLM}s at inference time.
\newblock In \emph{Proceedings of the 40th International Conference on Machine
  Learning (ICML)}, volume 202 of \emph{PMLR}, pages 22137--22176, 2023.
\newblock URL \url{https://arxiv.org/abs/2310.17157}.

\bibitem[Mallen et~al.(2023)Mallen, Asai, Zhong, Das, Khashabi, and
  Hajishirzi]{mallen2023trust}
Alex Mallen, Akari Asai, Victor Zhong, Rajarshi Das, Daniel Khashabi, and
  Hannaneh Hajishirzi.
\newblock When not to trust language models: Investigating effectiveness of
  parametric and non-parametric memories.
\newblock In \emph{Proceedings of the 61st Annual Meeting of the Association
  for Computational Linguistics (Volume 1: Long Papers)}, pages 9802--9822.
  Association for Computational Linguistics, 2023.
\newblock \doi{10.18653/v1/2023.acl-long.546}.
\newblock URL \url{https://aclanthology.org/2023.acl-long.546/}.

\bibitem[Manakul et~al.(2023)Manakul, Liusie, and
  Gales]{manakul2023selfcheckgpt}
Potsawee Manakul, Adian Liusie, and Mark J.~F. Gales.
\newblock {SelfCheckGPT}: Zero-resource black-box hallucination detection for
  generative large language models.
\newblock In \emph{Proceedings of the 2023 Conference on Empirical Methods in
  Natural Language Processing (EMNLP)}. Association for Computational
  Linguistics, 2023.
\newblock URL \url{https://arxiv.org/abs/2303.08896}.

\bibitem[Moreno~Cencerrado et~al.(2025)Moreno~Cencerrado, Padr{\'e}s~Masdemont,
  Gonzalvez~Hawthorne, Africa, and Pacchiardi]{cencerrado2025noanswer}
Iv{\'a}n~Vicente Moreno~Cencerrado, Arnau Padr{\'e}s~Masdemont, Anton
  Gonzalvez~Hawthorne, David~Demitri Africa, and Lorenzo Pacchiardi.
\newblock No answer needed: Predicting {LLM} answer accuracy from question-only
  linear probes, 2025.
\newblock URL \url{https://arxiv.org/abs/2509.10625}.
\newblock Accepted (poster), Principled Design for Trustworthy AI Workshop,
  ICLR 2026.

\bibitem[Muralidharan et~al.(2024)Muralidharan, Sreenivas, Joshi, Chochowski,
  Patwary, Shoeybi, Catanzaro, Kautz, and Molchanov]{muralidharan2024minitron}
Saurav Muralidharan, Sharath~Turuvekere Sreenivas, Raviraj Joshi, Marcin
  Chochowski, Mostofa Patwary, Mohammad Shoeybi, Bryan Catanzaro, Jan Kautz,
  and Pavlo Molchanov.
\newblock Compact language models via pruning and knowledge distillation.
\newblock In \emph{Advances in Neural Information Processing Systems
  (NeurIPS)}, volume~37, 2024.
\newblock URL \url{https://openreview.net/forum?id=9U0nLnNMJ7}.

\bibitem[nostalgebraist(2020)]{nostalgebraist2020logitlens}
nostalgebraist.
\newblock interpreting {GPT}: the logit lens.
\newblock LessWrong, 2020.
\newblock URL
  \url{https://www.lesswrong.com/posts/AcKRB8wDpdaN6v6ru/interpreting-gpt-the-logit-lens}.
\newblock Published August 31, 2020.

\bibitem[Obeso et~al.(2025)Obeso, Arditi, Ferrando, Freeman, Holmes, and
  Nanda]{obeso2025realtime}
Oscar Obeso, Andy Arditi, Javier Ferrando, Joshua Freeman, Cameron Holmes, and
  Neel Nanda.
\newblock Real-time detection of hallucinated entities in long-form generation,
  2025.
\newblock URL \url{https://arxiv.org/abs/2509.03531}.

\bibitem[Ociepa et~al.(2025{\natexlab{a}})Ociepa, Flis, Kinas, Wr{\'o}bel, and
  Gwo{\'z}dziej]{ociepa2025bielikv3small}
Krzysztof Ociepa, {\L}ukasz Flis, Remigiusz Kinas, Krzysztof Wr{\'o}bel, and
  Adrian Gwo{\'z}dziej.
\newblock Bielik v3 small: Technical report, 2025{\natexlab{a}}.
\newblock URL \url{https://arxiv.org/abs/2505.02550}.

\bibitem[Ociepa et~al.(2025{\natexlab{b}})Ociepa, Flis, Wr{\'o}bel,
  Gwo{\'z}dziej, and Kinas]{ociepa2025bielik11bv2}
Krzysztof Ociepa, {\L}ukasz Flis, Krzysztof Wr{\'o}bel, Adrian Gwo{\'z}dziej,
  and Remigiusz Kinas.
\newblock Bielik 11b v2 technical report, 2025{\natexlab{b}}.
\newblock URL \url{https://arxiv.org/abs/2505.02410}.

\bibitem[Orgad et~al.(2025)Orgad, Toker, Gekhman, Reichart, Szpektor, Kotek,
  and Belinkov]{orgad2025llmsknow}
Hadas Orgad, Michael Toker, Zorik Gekhman, Roi Reichart, Idan Szpektor, Hadas
  Kotek, and Yonatan Belinkov.
\newblock {LLM}s know more than they show: On the intrinsic representation of
  {LLM} hallucinations.
\newblock In \emph{International Conference on Learning Representations
  (ICLR)}, 2025.
\newblock URL \url{https://openreview.net/forum?id=KRnsX5Em3W}.

\bibitem[Park et~al.(2025)Park, Du, Yeh, Wang, and Li]{park2025tsv}
Seongheon Park, Xuefeng Du, Min-Hsuan Yeh, Haobo Wang, and Yixuan Li.
\newblock Steer {LLM} latents for hallucination detection, 2025.
\newblock URL \url{https://arxiv.org/abs/2503.01917}.
\newblock International Conference on Machine Learning (ICML) 2025.

\bibitem[Roy and Vetterli(2007)]{roy2007effective}
Olivier Roy and Martin Vetterli.
\newblock The effective rank: {A} measure of effective dimensionality.
\newblock In \emph{15th European Signal Processing Conference (EUSIPCO)}, pages
  606--610, Pozna{\'n}, Poland, 2007.
\newblock URL
  \url{https://www.eurasip.org/Proceedings/Eusipco/Eusipco2007/Papers/a5p-h05.pdf}.

\bibitem[Sriramanan et~al.(2024)Sriramanan, Bharti, Sadasivan, Saha,
  Kattakinda, and Feizi]{sriramanan2024llmcheck}
Gaurang Sriramanan, Siddhant Bharti, Vinu~Sankar Sadasivan, Shoumik Saha,
  Priyatham Kattakinda, and Soheil Feizi.
\newblock {LLM}-check: Investigating detection of hallucinations in large
  language models.
\newblock In \emph{Advances in Neural Information Processing Systems
  (NeurIPS)}, volume~37, pages 34188--34216, 2024.
\newblock URL
  \url{https://proceedings.neurips.cc/paper_files/paper/2024/hash/3c1e1fdf305195cd620c118aaa9717ad-Abstract-Conference.html}.

\bibitem[Su et~al.(2024)Su, Wang, Ai, Hu, Wu, Zhou, and Liu]{su2024mind}
Weihang Su, Changyue Wang, Qingyao Ai, Yiran Hu, Zhijing Wu, Yujia Zhou, and
  Yiqun Liu.
\newblock {MIND}: Unsupervised real-time hallucination detection based on the
  internal states of large language models.
\newblock In \emph{Findings of the Association for Computational Linguistics:
  ACL 2024}, pages 14379--14391. Association for Computational Linguistics,
  2024.
\newblock URL \url{https://aclanthology.org/2024.findings-acl.854/}.

\bibitem[Sun et~al.(2024)Sun, Chen, Kolter, and Liu]{sun2024massive}
Mingjie Sun, Xinlei Chen, J.~Zico Kolter, and Zhuang Liu.
\newblock Massive activations in large language models.
\newblock \emph{Conference on Language Modeling (COLM)}, 2024.
\newblock URL \url{https://arxiv.org/abs/2402.17762}.

\bibitem[V{\'a}zquez et~al.(2025)V{\'a}zquez, Mickus, Zosa, Vahtola, Tiedemann,
  Sinha, Segonne, S{\'a}nchez-Vega, Raganato, Libovick{\'y}, Karlgren, Ji,
  Helcl, Guillou, de~Gibert, Bengoetxea, Attieh, and
  Apidianaki]{vazquez2025mushroom}
Ra{\'u}l V{\'a}zquez, Timothee Mickus, Elaine Zosa, Teemu Vahtola, J{\"o}rg
  Tiedemann, Aman Sinha, Vincent Segonne, Fernando S{\'a}nchez-Vega, Alessandro
  Raganato, Jind{\v{r}}ich Libovick{\'y}, Jussi Karlgren, Shaoxiong Ji,
  Jind{\v{r}}ich Helcl, Liane Guillou, Ona de~Gibert, Jaione Bengoetxea, Joseph
  Attieh, and Marianna Apidianaki.
\newblock {SemEval}-2025 task 3: {M}u-{SHROOM}, the multilingual shared task on
  hallucinations and related observable overgeneration mistakes.
\newblock In \emph{Proceedings of the 19th International Workshop on Semantic
  Evaluation (SemEval-2025)}. Association for Computational Linguistics, 2025.
\newblock URL \url{https://arxiv.org/abs/2504.11975}.

\bibitem[Wu et~al.(2025)Wu, Wang, Xiao, Peng, and Fu]{wu2025retrievalheads}
Wenhao Wu, Yizhong Wang, Guangxuan Xiao, Hao Peng, and Yao Fu.
\newblock Retrieval head mechanistically explains long-context factuality.
\newblock In \emph{International Conference on Learning Representations
  (ICLR)}, 2025.
\newblock URL \url{https://openreview.net/forum?id=EytBpUGB1Z}.
\newblock Oral presentation.

\bibitem[Yin et~al.(2023)Yin, Sun, Guo, Wu, Qiu, and Huang]{yin2023selfaware}
Zhangyue Yin, Qiushi Sun, Qipeng Guo, Jiawen Wu, Xipeng Qiu, and Xuanjing
  Huang.
\newblock Do large language models know what they don't know?
\newblock In \emph{Findings of the Association for Computational Linguistics:
  ACL 2023}, pages 8653--8665. Association for Computational Linguistics, 2023.
\newblock \doi{10.18653/v1/2023.findings-acl.551}.
\newblock URL \url{https://aclanthology.org/2023.findings-acl.551/}.

\end{thebibliography}
}

\appendix

\section{Non-Polish control family: Gemma-4}
\label{app:gemma}

To test whether the findings are Bielik-specific, we ran the unchanged Polish pipeline - identical entity lists, prompts, judges (strict: Claude Opus 4.8), measurement point, and statistics on two checkpoints of a non-Polish-centric family: \texttt{google/gemma-4-E4B-it} (MatFormer-heritage, 42 layers; paired with Bielik-4.5B) and \texttt{google/gemma-4-12b-it} (48 layers; paired with Bielik-11B), identified here by their Hugging Face model IDs. The only adaptation is architectural: the decoder lives under \texttt{model.language\_model} and dispersion hooks attach to the per-layer \texttt{mlp.act\_fn}, mirroring the Bielik hook path. One interpretive caveat is pre-registered in the analysis: the \known{} tier was designed to be famous \emph{for Bielik}, and Gemma's strict \known{} correctness is only 7--17\% (Table~\ref{tab:gemma}), so some ``known'' entities are genuinely borderline for Gemma; the fabricated-string-free \known{}-vs-\unknownreal{} contrast is therefore the primary familiarity claim for this family, with \known{}-vs-\fabricated{} reported alongside. Refusal counts here are entity-level, marker-based all-five-sample refusals per condition (not the LLM audit used for Bielik).

\begin{table}[h]
\centering
\footnotesize
\caption{Gemma-4 on the unchanged Polish pipeline (athletes behavioral axis; separability means over the four domains, prompt point). Bielik reference values: K-vs-F dispersion 0.95--1.00 per domain, probe transfer mean off-diagonal 0.92--0.99, and 2 refusals in 2{,}520 sampled answers.}
\label{tab:gemma}
\begin{tabular}{lcc}
\toprule
 & Gemma-4-E4B & Gemma-4-12B \\
\midrule
Strict \known{} correctness (athletes, 5/5) & 3/42 & 7/42 \\
Mean K-vs-F, dispersion / probe & 0.83 / 0.85 & 0.91 / 0.91 \\
Mean K-vs-UR (primary), dispersion / probe & 0.90 / 0.94 & 0.95 / 0.94 \\
Probe transfer, mean off-diagonal & 0.74 & 0.77 \\
All-refusal entities (athletes, of 42): \known{} / \unknownreal{} / \fabricated{} & 0 / 0 / 0 & 7 / 28 / 24 \\
\bottomrule
\end{tabular}
\end{table}

The familiarity signal replicates qualitatively - weaker, patchier, and, unlike Bielik, still rising with scale (mean K-vs-F dispersion 0.83$\rightarrow$0.91 from E4B to 12B; on this family the signal is \emph{not} at ceiling at the smaller scale). The per-domain structure is the informative part: cities and writers are Bielik-grade (K-vs-F dispersion 0.92--1.00), while athletes and musicians (the domains whose famous tier is most Polish-local) drop to 0.66--0.87, exactly where Gemma's actual familiarity with the \known{} tier is demonstrably low (strict correctness 3/42 and 7/42). The signal dilutes where genuine familiarity is low, as a frequency-based account predicts: it tracks what the model knows, not what the dataset labels as famous. Cross-domain transfer is correspondingly weaker (mean off-diagonal probe 0.74/0.77 vs.\ Bielik's 0.92--0.99), bounded by the weak source signal in the people domains rather than by the cities template. The familiarity-vs-reliability dissociation replicates (K-vs-UR probe 0.94 at both scales against strict correctness of 7--17\%) - with the abstention twist reported in Section~\ref{sec:discussion}: Gemma-4-12B, unlike every Bielik model and unlike Gemma-4-E4B, refuses condition-selectively in exactly the direction its internal familiarity signal points. The full multi-family treatment is deliberately out of scope here and reserved for follow-up work.

\end{document}